\documentclass[journal,twoside,web]{ieeecolor}
\usepackage{generic}
\usepackage{cite}
\usepackage{amsmath,amssymb,amsfonts}
\usepackage{algorithmic}
\usepackage{graphicx}
\usepackage{algorithm,algorithmic}
\usepackage{hyperref}
\hypersetup{hidelinks=true}
\usepackage{textcomp}

\usepackage{booktabs}
\usepackage{multirow}
\usepackage{soul}
\usepackage{url}
\usepackage{subcaption}
\usepackage{tabularx}
\usepackage{paralist}
\usepackage{flushend}
\usepackage{colortbl}

\graphicspath{{./img/}}

\usepackage[dvipsnames]{xcolor}

\newcommand{\ra}{\includegraphics[width=0.9em]{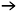}}
\newcommand{\rab}{\includegraphics[width=0.9em]{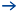}}

\newcommand{\HD}{$\text{HD}_{95}$}

\definecolor{mygray}{HTML}{ededed}

\DeclareRobustCommand{\fZ}{\raisebox{-0.3\height}{\includegraphics[width=1em]{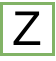}}}
\DeclareRobustCommand{\fAE}{\raisebox{-0.3\height}{\includegraphics[width=1.8em]{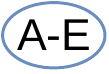}}}

\newcommand{\papertitle}{Deep Multimodal Fusion of Data with Heterogeneous Dimensionality via Projective Networks}

\def\BibTeX{{\rm B\kern-.05em{\sc i\kern-.025em b}\kern-.08em
    T\kern-.1667em\lower.7ex\hbox{E}\kern-.125emX}}
\begin{document}

\title{\papertitle}
\author{%
    José Morano,
    Guilherme Aresta,
    Christoph Grechenig,
    Ursula Schmidt-Erfurth,
    and Hrvoje Bogunović
    \thanks{%
        This work was supported in part by the Christian Doppler Research Association, Austrian Federal Ministry for Digital and Economic Affairs, the National Foundation for Research, Technology and Development, and Heidelberg Engineering.
    }
    \thanks{%
        J. Morano, G. Aresta, and H. Bogunovi\'c are with the Christian Doppler Laboratory for Artificial Intelligence in Retina, Department of Ophthalmology and Optometry, Medical University of Vienna, Austria (e-mail: \{jose.moranosanchez, hrvoje.bogunovic\}@meduniwien.ac.at).
        C. Grechenig, and U. Schmidt-Erfurth, are with the Department of Ophthalmology and Optometry, Medical University of Vienna, Austria.
    }
}

\maketitle

\begin{abstract}
    The use of multimodal imaging has led to significant improvements in the diagnosis and treatment of many diseases.
    Similar to clinical practice, some works have demonstrated the benefits of multimodal fusion for automatic segmentation and classification using deep learning-based methods.
    However, current segmentation methods are limited to fusion of modalities with the same dimensionality (e.g., 3D+3D, 2D+2D), which is not always possible, and the fusion strategies implemented by classification methods are incompatible with localization tasks.
    In this work, we propose a novel deep learning-based framework for the fusion of multimodal data with heterogeneous dimensionality (e.g., 3D+2D) that is compatible with localization tasks.
    The proposed framework extracts the features of the different modalities and projects them into the common feature subspace.
    The projected features are then fused and further processed to obtain the final prediction.
    The framework was validated on the following tasks: segmentation of geographic atrophy (GA), a late-stage manifestation of age-related macular degeneration, and segmentation of retinal blood vessels (RBV) in multimodal retinal imaging.
    Our results show that the proposed method outperforms the state-of-the-art monomodal methods on GA and RBV segmentation by up to 3.10\% and 4.64\% Dice, respectively.
\end{abstract}


\begin{figure}[htb]
\centering
\includegraphics[width=0.99\linewidth]{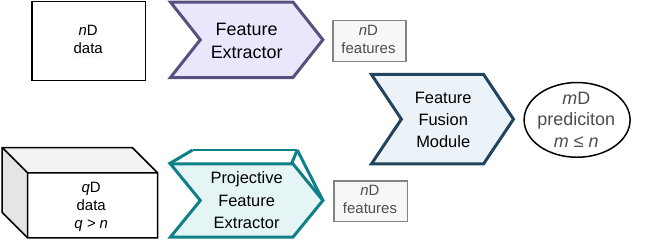}
    \caption{
        Our proposed framework
        defines a novel fusion approach that extracts and projects the features of all modalities into the feature space of the modality with the lowest dimensionality ($n$), so that
        they can be employed for localization tasks in the common ($n$-dimensional, $n$D) subspace.
    }
\label{fig:idea}
\end{figure}

\section{Introduction}

\IEEEPARstart{U}{sing}
complementary information provided by different imaging modalities, commonly referred to as \emph{multimodal imaging}, has led to significant improvements in the diagnosis and monitoring of several diseases~\cite{Cai_IEEEA_2019,Zhang_IF_2020,Boehm_Nature_2021,Azam_CBM_2022,Ambrosini_EJR_2012,Nasir_ISWA_2023,Kanski_Elsevier_2011,Wu_GAMMA_arXiv_2022}.
For instance, magnetic resonance imaging (MRI), which provides high-resolution images of soft tissue, can be combined with computed tomography (CT), which better captures information from hard tissue with little distortion~\cite{Azam_CBM_2022}, for the diagnosis of brain diseases.
Likewise, X-ray and CT can be used together for a more accurate diagnosis of lung nodules~\cite{Ambrosini_EJR_2012} and breast cancer~\cite{Nasir_ISWA_2023}.
In ophthalmology, optical coherence tomography (OCT)~\cite{Huang_OCT_1991}, a non-invasive 3-dimensional (3D) imaging technique that allows obtaining cross-sectional images (B-Scans) of the retina, is often combined with 2D fundus imaging, such as color fundus photography (CFP), confocal scanning laser ophthalmoscopy (SLO)~\cite{webb1987confocal} or fundus autofluorescence (FAF)~\cite{delori1995vivo}, for the diagnosis and monitoring of diseases such as age-related macular degeneration (AMD)~\cite{Guymer_CEO_2020,riedl_effect_2022,bui_fundus_2022} or diabetic retinopathy~\cite{Porwal_MIA_2020}.

Reliable quantitative analysis of medical images requires precise segmentation of the anatomical structures or lesions of interest~\cite{Liu_LDH_2019}.
However, manual segmentation is time-consuming, subject to interpretation, and prone to human error, increasing costs and affecting reproducibility.
Deep learning has shown great potential to automate this process, and is currently the state of the art in many (semi-)automatic medical segmentation tasks~\cite{Menze_BRATS_TMI_2015,Liu_LDH_2019,Wu_GAMMA_arXiv_2022}

Similar to clinical practice, automated image segmentation models can also benefit from the use of multimodal data, usually referred to as \emph{multimodal image fusion} (MIF)~\cite{Dolz_CSI_2019,zhang2021brain,Azam_CBM_2022,zhou2019review}.
In particular, previous work has shown that MIF can improve segmentation performance with respect to monomodal baselines~\cite{Dolz_CSI_2019,zhang2021brain,Azam_CBM_2022,zhou2019review}.
Moreover, MIF can be useful in scenarios with limited annotated data, as it increases the amount of information available to the model without the need for additional annotations.
This is especially relevant for medical applications of deep learning, since deep learning models usually require large amounts of data to achieve good performance, and medical data is often scarce and expensive to obtain.
To the best of our knowledge, existing segmentation methods have only addressed the fusion of data of the same dimensionality (either 2D or 3D), and none of them can be directly applied to multimodal data with \emph{heterogeneous} dimensionality (e.g., 3D+2D).
Similarly, the current state-of-the-art classification methods that do combine data of different dimensionality~\cite{Wu_GAMMA_arXiv_2022} are unsuitable for localization tasks (e.g., segmentation and detection) due to the fusion process they employ, which first projects the features onto 1D vectors (thereby discarding the spatial information) and then concatenates them before being fed into the final fully connected (FC) layers.

Furthermore, there are segmentation tasks where the target output has a different dimensionality than the input.
The most common of these scenarios is 3D-to-2D (3D\ra2D), where the input is a 3D volume and the segmentation is performed on a 2D projection of the volume.
This occurs, for example, in the segmentation of geographic atrophy (GA), a late stage of AMD, in OCT or the segmentation of retinal blood vessels (RBV) in OCT and OCT angiography (OCT-A), where segmentation is performed on the 2D OCT projection~\cite{Lachinov_MICCAI_2021,Li_IPN_TMI_2020}.
In recent years, several methods have been proposed for this type of tasks using \emph{monomodal} inputs~\cite{Lachinov_MICCAI_2021,Li_IPN_TMI_2020,Li_IPNv2_arXiv_2020}.
However, the application of these methods to \emph{multimodal} data has not yet been explored, as no suitable fusion method is available.

In this work, we propose a novel framework for fusing multimodal data of heterogeneous dimensionality (Fig.~\ref{fig:idea}) that, unlike existing methods, is compatible with localization tasks.
The framework defines a novel fusion approach that extracts and projects the features of all modalities into a common feature subspace, enabling their use for localization tasks.
To assess its potential, we propose and thoroughly validate two different fusion approaches based on it: Late Fusion and Multiscale Fusion.
The proposed approaches are validated on two clinically relevant tasks: segmentation of geographic atrophy (GA) and segmentation of retinal blood vessels (RBV) in multimodal retinal imaging.
Our results show that our method significantly outperforms the state-of-the-art monomodal methods on both tasks.

\subsection{Related Work}
\subsubsection{Multimodal Image Fusion}

In recent years, due to the increasing availability of multimodal image data, MIF has become a very active research topic~\cite{zhou2019review,yadav2020image,tan2020multimodal,muhammad2021comprehensive,zhang2021image,hermessi2021multimodal,Azam_CBM_2022,stahlschmidt2022multimodal}.
MIF approaches can be divided into three main categories, depending on the part of the model where the fusion occurs: Input-Level (IL), Output-Level (OL), and Layer-Level (LL)~\cite{zhou2019review}.

\emph{IL fusion} operates directly on the raw pixels/voxels of the images, combining them into a single image that is then fed to the model~\cite{bhavana2015multi,shabanzade2017combination,singh2018ripplet}.
Despite being straightforward to implement, this method cannot be directly used for fusing data of heterogeneous dimensionality (e.g., 3D volumes and 2D images).
Moreover, previous works have shown that IL fusion may be inefficient~\cite{liu2016multispectral,ramachandram2017deep} and insufficient to fully exploit multimodal information~\cite{liu2016multispectral,ramachandram2017deep,Dolz_CSI_2019}.

In \emph{OL fusion}, the images from the different modalities are fed to two or more independent models, and the resulting predictions are combined at the end.
For example, for brain tumor segmentation, Kamnitsas, \textit{et al.}~\cite{kamnitsas2018ensembles} trained three networks separately and then averaged the confidence of each network.
The final segmentation is obtained by assigning each voxel with the highest confidence, which depends on the majority voting.
OL fusion strategy is more flexible than IL fusion, but it requires training multiple models and it does not allow to deeply exploit the relationship between the different modalities in the learning process.

Lastly, in \emph{LL fusion}, the images from the different modalities are fed to the model separately, but their representations are combined at some point and processed jointly to obtain the final result.
This strategy does not have any of the limitations of IL and OL fusion, since it can be used with heterogeneous data and it allows to deeply exploit the relationship between the different modalities, but its performance greatly depends on the way the representations are combined.
For classification, the most common approach is to concatenate the 1D embeddings of the images from the different modalities and feed them to one or more FC layers~\mbox{\cite{Wu_GAMMA_arXiv_2022,wang2022adversarial,sharif2022m3btcnet}}.
Wang \textit{et al.}~\cite{wang2022adversarial} proposed a fusion method for skin lesion classification that uses a single encoder for all modalities and exploits correlated and complementary multimodal information by using adversarial learning and attention.
Sharif \textit{et al.}~\mbox{\cite{sharif2022m3btcnet}} proposed a multimodal classification method for MRI images that optimizes the features extracted from a CNN using evolutionary algorithms.
This fusion approach, however, is incompatible with localization tasks, such as segmentation, since the location information of the features is lost.
Other methods have demonstrated that fusing the representations from different modalities at multiple levels of the network can lead to better performance~\cite{Dolz_CSI_2019}.
For example, for multimodal segmentation of MRI volumes, Dolz \textit{et al.}~\cite{Dolz_CSI_2019} proposed an iterative fusion method consisting in a 2D fully convolutional neural network (FCNN) with multiple interconnected encoders, one per modality, and a common decoder that processes the features from all the encoders.
However, this method processes input volumes slice-wise and cannot be applied to modalities of different dimensionality.
Thus, despite the potential of the LL fusion strategy,
its application to images of heterogeneous dimensionality has not yet been explored.

\subsubsection{3D\rab2D Segmentation/Regression}

Several methods have been proposed for performing 2D segmentation from 3D input volumes~\cite{rafiei2018liver,bermejo2019sr,mckinley2019ensembles,mckinley2021simultaneous,Li_IPN_TMI_2020,Li_IPNv2_arXiv_2020,Lachinov_MICCAI_2021,Seebok_OR_2022,morano2023selfsupervised}.
Current state-of-the-art approaches are based on the use of 3D\ra2D FCNNs.
This type of network projects the 3D features from the input volume to 2D, and then processes them to obtain the final map.
The key differences between the methods lie in the way the features are projected to 2D and the network architecture.

For segmenting the liver in the central slice of a computed tomography (CT) volume, Rafiei \textit{et al.}~\cite{rafiei2018liver} propose an FCNN with a 3D encoder and a 2D decoder connected by skip connections.
Since 3D and 2D features cannot be directly concatenated due to their different dimensionality, only the features from the central slice of the 3D volume are used for the skip connections.
The same approach was used by Bermejo \textit{et al.}~\cite{bermejo2019sr} for segmenting paraseptal emphysema in chest CT volumes.
As an alternative to the network proposed in \cite{rafiei2018liver}, the authors propose a novel FCNN featuring Dense~\cite{huang2017densely} and ENet-like~\cite{paszke2016enet} blocks.
In both cases, however, the segmentation is only done for the central slice of the input volume, and the features from the other slices are only used in the bottleneck of the network.
This renders the method unsuitable for tasks where the segmentation is done on a projection of the input volume, where all the slices are equally important.
Moreover, these networks require the input volume to have a fixed size in the depth dimension, since the pooling layers consecutively reduce the depth dimension of the features to 1.

A network architecture based on both Dense and ENet-like blocks was also used by McKinley \textit{et al.}~\cite{mckinley2019ensembles,mckinley2021simultaneous} for segmenting brain tumors in MRI volumes.
However, in this case, 3D features are projected to 2D using 3D convolutional layers at the very beginning of the network.
Then, the resulting 2D features are processed by a U-Net-like~\cite{Ronneberger_MICCAI_2015} encoder-decoder architecture.
For RBV segmentation in OCT-A volumes, Li \textit{et al.}~\cite{Li_IPN_TMI_2020} proposed an image projection network (IPN) that reduces the features to the target dimensionality using unidirectional pooling layers in the encoder.
Later, Lachinov \textit{et al.}~\cite{Lachinov_MICCAI_2021} proposed a U-Net-like convolutional neural network (CNN) for 3D\ra2D segmentation that outperforms IPN and overcomes its main limitations, namely,
fixed input size, limited context, and lack of detail in the segmentation due to the absence of skip connections.
This approach~\cite{Lachinov_MICCAI_2021} was validated on the tasks of GA and RBV segmentation from OCT volumes.
IPN limitations were also later overcome by its second version~\cite{Li_IPNv2_arXiv_2020}, IPNv2, but its performance remained lower than that of Lachinov \textit{et al.} in various tasks~\cite{morano2023selfsupervised}.

In addition, there are works that explore the use of CNNs for 3D\ra2D regression, where Seeböck \textit{et al.}~\cite{Seebok_OR_2022} proposed ReSensNet, a novel CNN based on Residual 3D U-Net~\cite{Lee_Res3DUNet_arXiv_2017}, with a 3D encoder and a 2D decoder connected by 3D\ra2D blocks.
In particular, they applied ReSensNet to the task of estimating the retinal sensitivity from OCT volumes.
However, ReSensNet only works at concrete input resolutions, and it is applied pixel-wise.
In a recent work~\cite{morano2023selfsupervised}, we proposed a novel FCNN for 3D\ra2D segmentation based on ReSensNet~\cite{Seebok_OR_2022}, named Feature Projection Network (FPN), that solves these problems by using a novel type of 3D\ra2D block, Feature Projection Block (FPB), that replaces the 3D\ra2D Blocks of ReSensNet.
This FCNN surpasses the state of the art~\cite{Lachinov_MICCAI_2021,Li_IPN_TMI_2020,Li_IPNv2_arXiv_2020} in label-scarce scenarios for different tasks.

\subsection{Contributions}

The main contributions of our work are as follows:
\begin{itemize}
    \item We propose the first layer-level fusion framework for fusing multimodal data of heterogeneous dimensionality that is suitable for localization tasks, such as segmentation.
    \item Based on this framework, we propose two novel fusion approaches for 3D\ra2D segmentation: Late Fusion and Multiscale Fusion.
    Both approaches are implemented using a new FCNN architecture based on~\cite{morano2023selfsupervised}.
    \item The proposed fusion approaches significantly outperform state-of-the-art monomodal methods on two clinically relevant tasks (GA and RBV segmentation in multimodal retinal imaging) demonstrating the effectiveness of the proposed framework.
\end{itemize}

\section{Methods}%
\label{sec:methods}

\begin{figure}[tbp]
\centering
\includegraphics[width=0.99\linewidth]{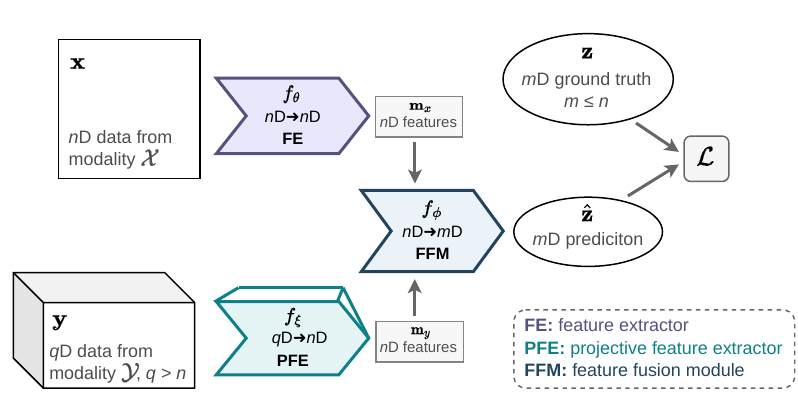}
    \caption{%
        Illustration of the proposed framework, consisting of 3 modules:
        feature extractor (FE), projective feature extractor (PFE), and feature fusion module (FFM).
        The FE extracts $n$D features from $n$D data, and the PFE extracts $n$D features from $q$D data by projecting them to the $n$D feature space.
        Then, the FFM processes the $n$D features extracted from the different modalities to obtain the final $m$D prediction, where $m \leq n$.
    }
\label{fig:framework}
\end{figure}

The proposed LL-based fusion framework is illustrated in Fig.~\ref{fig:framework}.
Although the framework supports an arbitrary number of modalities, for the sake of simplicity we describe it only for
two modalities: $\mathcal{X} \subset \mathbb{R}^{n}$ and $\mathcal{Y} \subset \mathbb{R}^{q}$, where $q > n$.
The framework consists of
the following fully convolutional modules:
\begin{inparaenum}[(1)]
\item \emph{feature extractor} (FE), that extracts $n$-dimensional ($n$D) features from $n$D data;
\item \emph{projective feature extractor} (PFE), that extracts $n$D features from $q$D data by projecting them to the $n$D feature space;
and
\item \emph{feature fusion module} (FFM), that combines and processes the $n$D features extracted from the different input data to produce the final prediction.
In this way, the final prediction can have any dimensionality $m \leq n$; e.g., it can be a classification label, a segmentation mask, or a regression value.
The framework can be extended to any number of modalities by adding more FEs and PFEs, that may or may not share parameters among themselves.
\end{inparaenum}

Specifically, let $\mathbf{x} \in \mathcal{X}$
and $\mathbf{y} \in \mathcal{Y}$
be two images from different modalities
and $\mathbf{z} \in \mathbb{R}^{m}$, where $m\leq n$, their corresponding common ground truth.
We extract features $\mathbf{m}_x = f_{\theta}(\mathbf{x})$ and $\mathbf{m}_y = f_{\xi}(\mathbf{y})$ from $\mathbf{x}$ and $\mathbf{y}$ using FE $f_{\theta}$ and PFE $f_{\xi}$.
The PFE $f_{\xi}$ projects $q$-dimensional features $\mathbf{m}_y$ to the $n$-dimensional feature space, so that $\mathbf{m}_y \in \mathbb{R}^{n}$.
In this way, the feature space of $\mathbf{m}_x$ and $\mathbf{m}_y$ is of the same dimensionality, and thus they can be processed simultaneously using the same FFM.
Lastly, we obtain the
final
prediction $\mathbf{z} = f_{\phi}(\mathbf{m}_x, \mathbf{m}_y)$ using FFM $f_{\phi}$, that processes $\mathbf{m}_x$ and $\mathbf{m}_y$.
The $f_{\phi}$ defines how features $\mathbf{m}_x$ and $\mathbf{m}_y$ are fused
and processed to obtain the final prediction
$\hat{\mathbf{z}}$.
The different modules constitute a joint model $f_{\phi, \theta, \xi}(\mathbf{x}, \mathbf{y}) = \hat{\mathbf{z}}$ that can be trained end-to-end using any loss function $\mathcal{L(\hat{\mathbf{z}}, \mathbf{z})}$.

To evaluate the effectiveness of our multimodal fusion framework, we propose two different fusion approaches for 2D segmentation in 3D and 2D modalities that are based on this framework: Late Fusion and Multiscale Fusion (Subsection~\ref{sec:fusion_approaches}).
In addition, we propose a novel FCNN architecture that, with some modifications, can be used to implement both approaches (Subsection~\ref{sec:network_architecture}).

\subsection{Fusion Approaches}%
\label{sec:fusion_approaches}

\subsubsection{Late Fusion}%
\label{sec:late_fusion}

\begin{figure}[htbp]
    \centering
    \includegraphics[width=0.99\linewidth]{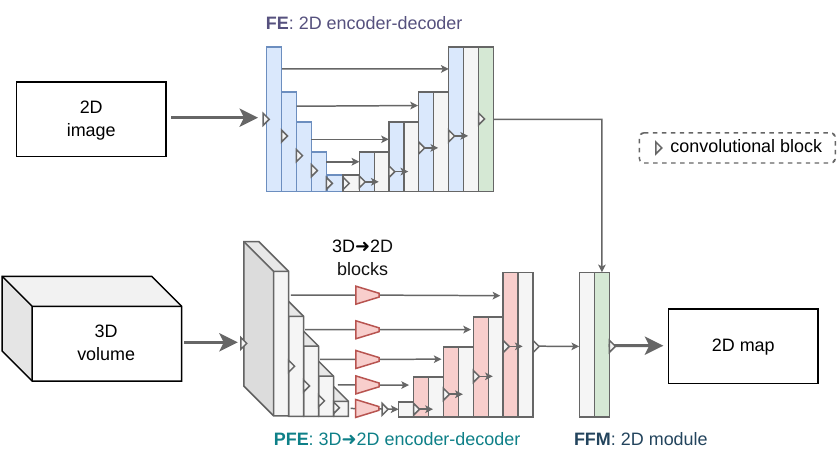}
    \caption{%
        Illustration of the Late Fusion approach.
        The features extracted from different encoder-decoder modules for each modality are concatenated and processed by a simple convolutional block.
    }
    \label{fig:late}
\end{figure}

In this approach (Fig.~\ref{fig:late}), both FE and PFE have encoder-decoder architectures that process the corresponding input data and output a set of feature maps with the same size as the input.
To obtain the final prediction, these feature maps are concatenated and processed by a convolutional block (here, the FFM).
This approach is similar to the fusion approach commonly used for classification~\cite{Wu_GAMMA_arXiv_2022}, but with the important difference that, in this case, the concatenated features are $n$D, and not necessarily 1D.

\subsubsection{Multiscale Fusion}%
\label{sec:multiscale_fusion}

\begin{figure}[htbp]
    \centering
    \includegraphics[width=0.99\linewidth]{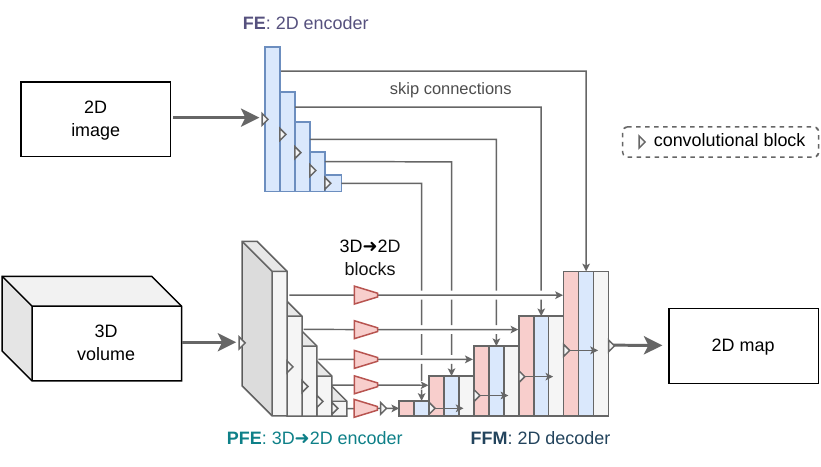}
    \caption{%
        Illustration of the
        Multiscale Fusion approach.
        The features extracted from different encoders for each modality are fused at multiple scales and processed by a single decoder.
    }
    \label{fig:multiscale}
\end{figure}

The proposed Multiscale Fusion approach (Fig.~\ref{fig:multiscale}), inspired by~\cite{Dolz_CSI_2019}, consists of two different fully convolutional encoders (acting as FE and PFE), one for each modality, and a single decoder (FFM).
Both encoders are connected to a common decoder through skip connections at multiple levels.

\subsection{Network Architecture}%
\label{sec:network_architecture}

\begin{figure*}[tbhp]
    \centering
    \includegraphics[width=\linewidth]{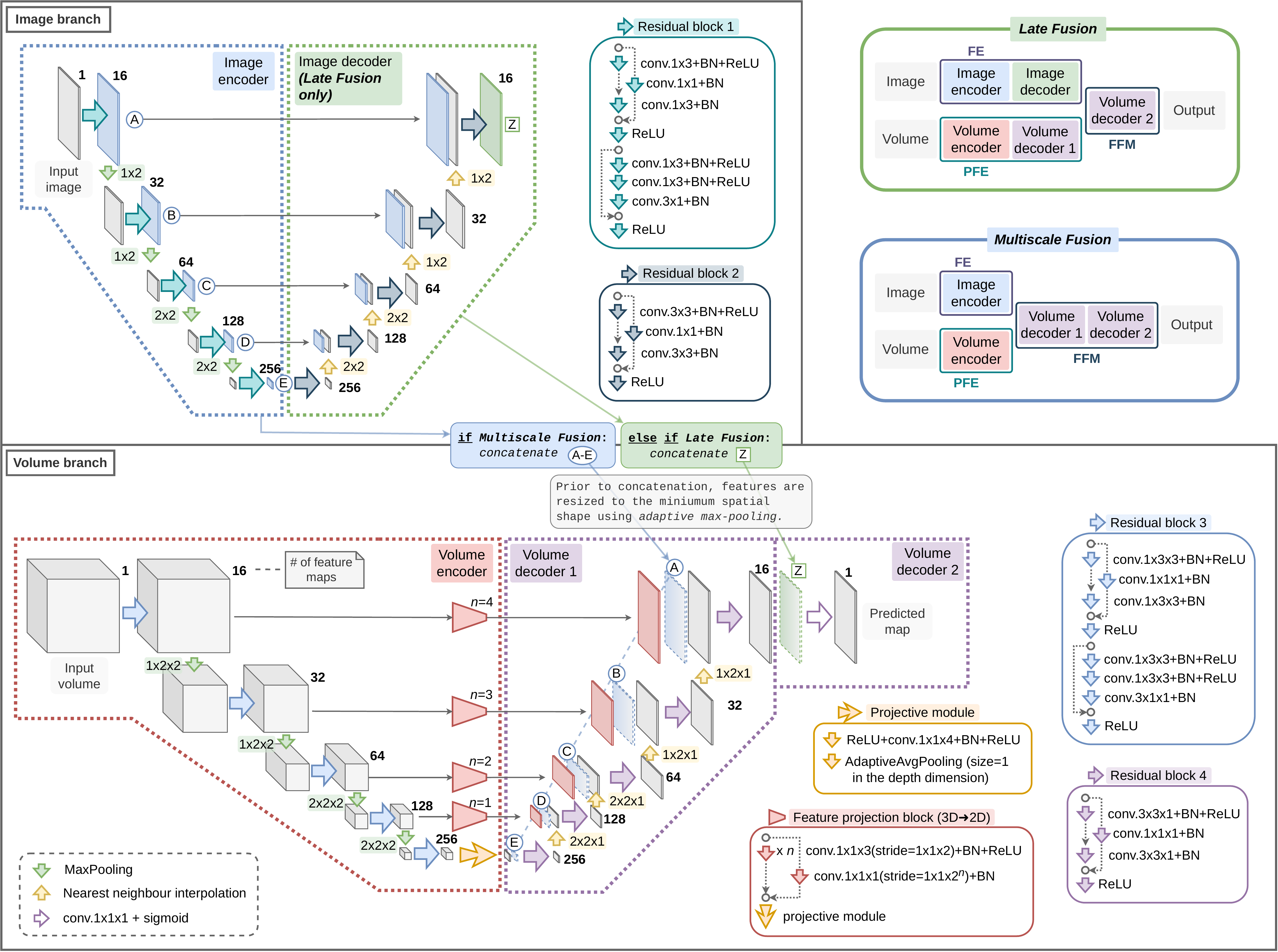}
    \caption{%
        Proposed FCNN for multimodal fusion.
        For Late Fusion, the Image branch is an encoder-decoder architecture, and its output (\fZ) is concatenated with the output of the Volume branch before the final convolution.
        For Multiscale Fusion, only the encoder of the Image branch is used, and its features at different scales (\fAE) are concatenated with the features of the encoder from the Volume branch and fed to a common decoder.
        Before concatenation, all features are resized to the minimum feature size using adaptive max-pooling.
        The number of feature maps is indicated in bold.
    }
    \label{fig:architecture}
\end{figure*}

We propose an FCNN architecture (Fig.~\ref{fig:architecture}) based on \cite{morano2023selfsupervised} that can be adapted to implement both Late Fusion and Multiscale Fusion approaches.
The architecture is composed of two main branches: Image branch and Volume branch.

The Image branch is an FCNN composed of a 2D encoder and a 2D decoder connected by skip connections.
Both are composed of 5 residual blocks of different type.
In the encoder, each residual block has 8 2D convolutional layers; in the decoder, each block has 4 layers.
The decoder outputs 16 feature maps with the same spatial resolution as the input images (\fZ, in Fig.~\ref{fig:architecture}).

The Volume branch is an FCNN composed of a 3D encoder and a 2D decoder connected by 3D\ra2D FPBs.
The encoder and the decoder are composed of different types of residual blocks.
The 3D encoder is composed of 5 residual blocks with 8 convolutional layers with 3D kernels, and the 2D decoder is composed of 5 residual blocks with 4 convolutional layers with 2D kernels.
FPBs project 3D features to the 2D feature space.
They are composed of a variable number of convolutions followed by a depth-wise adaptive average pooling of size 1.
The outputs of the FPBs are connected to the decoder through skip connections.
At the end of the decoder, a $1 \times 1 \times 1$ convolution and a sigmoid function are applied to obtain the final 2D map.

The way these branches are connected depends on the fusion approach.
In Late Fusion, the output of the Image decoder (\fZ, in Fig.~\ref{fig:architecture}) is concatenated with the output of the Volume decoder before the last $1 \times 1 \times 1$ convolution.
In Multiscale Fusion, the outputs of the Image encoder (\fAE, in Fig.~\ref{fig:architecture}) and the Volume encoder (after the projective modules) are concatenated and fed to the Volume decoder.
Thus, the Image decoder is not used in this case.

Since the number of pixels of the 2D image and the 3D volume in their common dimensions may not match, and therefore those of their corresponding features, we resize all the features to be concatenated to the minimum feature size using adaptive max pooling.

\section{Experimental Setup}%
\label{sec:experimental_setup}

The proposed multimodal fusion approaches were evaluated on two clinically relevant tasks: segmentation of GA and RBV in multimodal 3D and 2D retinal images.
For the experiments, we used two different in-house datasets (Vienna Clinical Trial Center at the
Department of Ophthalmology and Optometry, Medical University of Vienna) with multimodal data and manual segmentation maps.
The analysis adhered to the tenets of the Declaration of Helsinki, and approval was obtained by the Ethics Committee of the Medical University of Vienna (Nr 1246/2016).

\subsection{Clinical Background}\label{sec:clinical_background}

\emph{Multimodal imaging} plays an important role in the diagnosis and treatment of many retinal diseases.
Three of the most commonly used retinal imaging modalities are OCT, SLO, and FAF.
All three are non-invasive and can be used to image the retina in vivo.
OCT provides high-resolution cross-sectional images of the retina by measuring the echo time delay and intensity of backscattered light from the retina~\cite{Huang_OCT_1991}.
Typically, multiple B-scans are acquired in a raster pattern to obtain a 3D volume.
SLO is a 2D imaging modality based on the principle of confocal imaging that captures the backscattered light from the retina in a raster pattern~\cite{webb1987confocal}.
FAF, also 2D, captures the natural and pathological fluorescence occurring in the retina by illuminating it with a blue light source and capturing the emitted light in the green wavelength range~\cite{delori1995vivo}.

\begin{figure*}[tbhp]
\centering
\begin{tabular}{ccccc}
\includegraphics[height=0.17\textwidth]{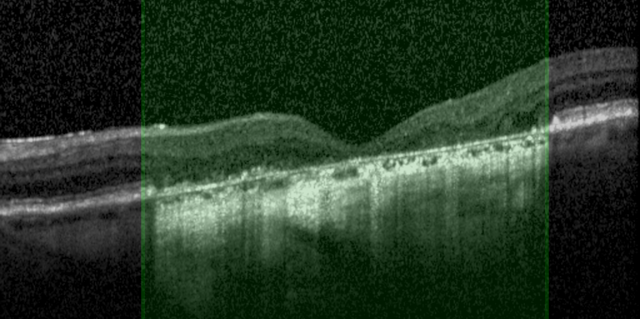}
& \includegraphics[height=0.17\textwidth]{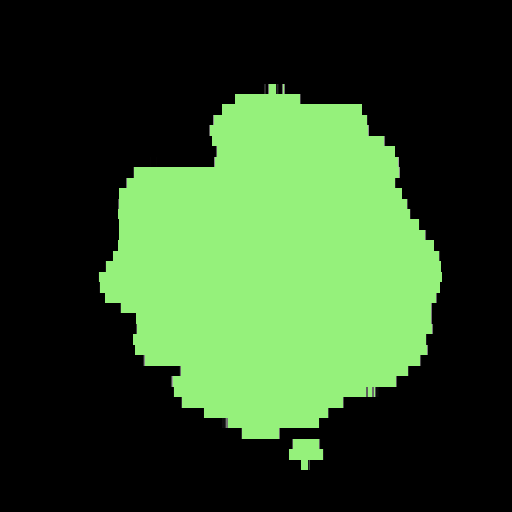}
& \includegraphics[height=0.17\textwidth]{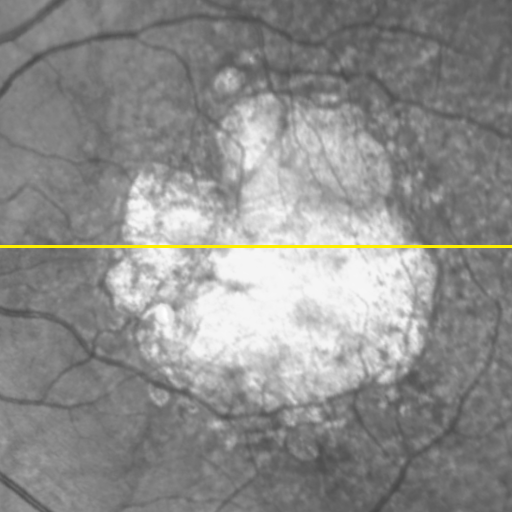}
& \includegraphics[height=0.17\textwidth]{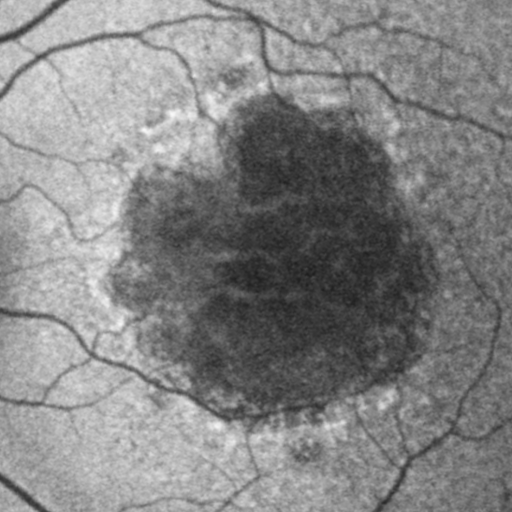} \\

\includegraphics[height=0.17\textwidth]{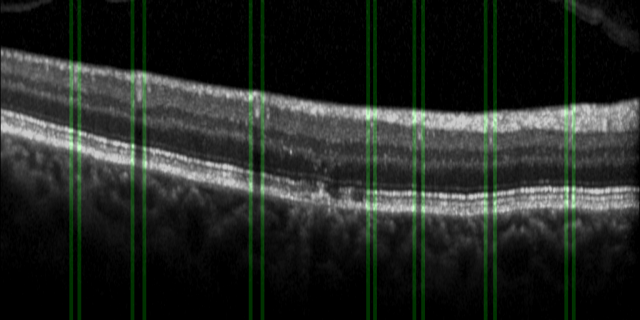}
& \includegraphics[height=0.17\textwidth]{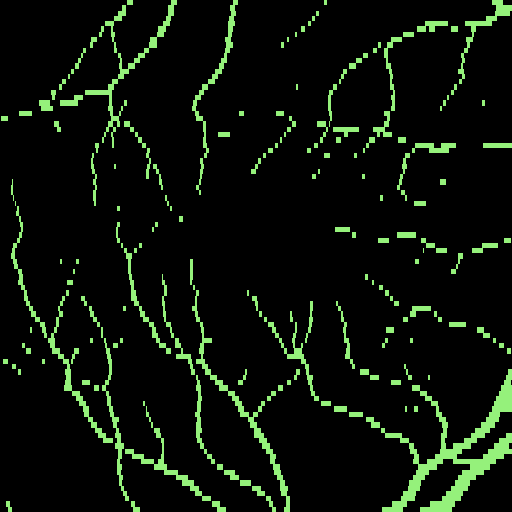}
& \includegraphics[height=0.17\textwidth]{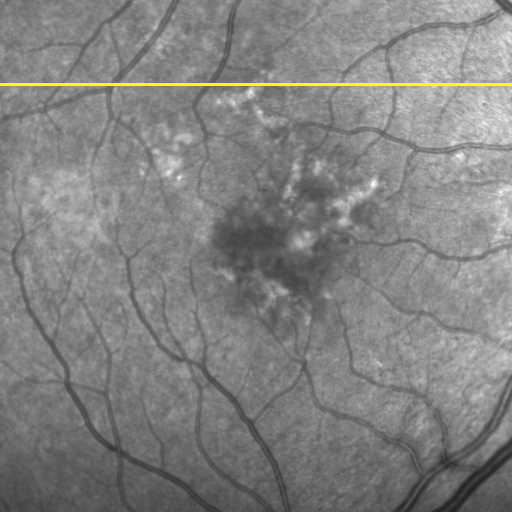}
&
\end{tabular}
\caption{%
From left to right: OCT slice (B-Scan) with the corresponding
reference
annotations overlaid in green, reference en-face map, SLO with the location of the B-Scan indicated in yellow, and FAF.
Top: geographic atrophy.
Bottom: retinal blood vessels.
}%
\label{fig:clinical_background}
\end{figure*}

\emph{GA} is an advanced form of AMD
that corresponds to a progressive loss of retinal photoreceptors and leads to irreversible visual impairment~\cite{Kanski_Elsevier_2011}.
It is typically assessed with OCT and/or FAF imaging modalities~\cite{Bui_Eye_2021,Wei_Eye_2023}.
In OCT, GA is characterized by the loss of retinal pigment epithelium (RPE) tissue, accompanied by the contrast enhancement of the signal below the retina (see Fig.~\ref{fig:clinical_background}, top).
In FAF, GA is characterized by the loss of RPE autofluorescence~\cite{SchmitzValckenberg_O_2016}, so it appears darker than the surrounding areas.
Since GA essentially denotes a loss of tissue, it does not have ``thickness''.
Thus, in both cases, the GA lesion is delineated as a 2D en-face area.
Also, GA frequently appears brighter than the surrounding areas in SLO images due to its higher reflectance.

\emph{RBV} provide oxygen and nutrition to the retinal tissues.
In OCT B-Scans, they can be detected
as bright large dots
dropping shadows on the underlying retinal structures (see Fig.~\ref{fig:clinical_background}, bottom).
In SLO images, they typically appear darker than the surrounding tissue due to their lower reflectance~\cite{Kromer_JMBE_2016}.

\subsection{Geographic Atrophy Segmentation}\label{sec:ga_segmentation}

\subsubsection{Data}\label{sec:ga:data}
For the experiments, we used an in-house dataset consisting of 967 OCT volumetric scans from 100 patients (184 eyes) with GA.
All samples come from a clinical study on natural GA progression.
The scans were acquired using a Spectralis OCT device (Heidelberg Engineering, DE).
The dataset includes OCT B-Scans, SLO and FAF images.
All samples have GA en-face masks annotated by a retinal expert on FAF images.
For evaluation purposes, additional GA annotation was performed by retinal experts directly on the OCT for 35 samples from different patients.
OCT and SLO images were automatically co-registered by the imaging device, while FAF images were registered with SLO using an in-house image registration pipeline based on aligning retinal vessel segmentation~\cite{arikan2019deep}.
Once registered, FAF and SLO images were cropped and resized to the same area and resolution as the OCT en-face projection.
The approximate volume of each OCT is $6 \times 6 \times 1.92$ mm\textsuperscript{3} (en-face height $\times$ en-face width $\times$ depth), with a size of $49 \times 1024 \times 496$ voxels.
After the registration, cropping and resizing, SLO and FAF images have an en-face size of $49 \times 1024$ pixels.

\subsubsection{Experiments}%
\label{sec:ga_segmentation_experiments}

\paragraph{Ablation study}
We performed an ablation study to assess the impact of the fusion approaches (OCT+FAF, OCT+SLO) by comparing their performance with that of the Image branch (Image-br) of our architecture (see Fig.~\ref{fig:architecture}) trained solely with FAF/SLO images, and with the Volume branch (Volume-br) (see Fig.~\ref{fig:architecture}) trained solely with OCT.

\paragraph{State of the art comparison}
We performed a comparison with 2D U-Net~\cite{Ronneberger_MICCAI_2015} trained on FAF images
and three other state-of-the-art 3D\ra2D methods trained on OCT: Lachinov \textit{et al.}~\cite{Lachinov_MICCAI_2021}, ReSensNet~\cite{Seebok_OR_2022}, and our previous work, FPN~\cite{morano2023selfsupervised}.
Since our Volume branch coincides with the FPN architecture, we will refer to this baseline method as ``FPN/Volume-br''.

\paragraph{Data efficiency}
To evaluate the performance of the proposed approach with a limited number of labeled samples, we performed all experiments with different percentages (pcts.) of the training data: 10\%, 20\% and 100\%.

\paragraph{Model robustness}
The robustness of the multimodal models with respect to the best performing monomodal models was assessed by means of two extra evaluations.
First, we evaluated their performance using target OCT-based annotations instead of FAF-based.
Second, under noisy data.
In particular, we applied \textit{cutout} (randomly masking square regions of the data) to the OCT inputs.
The latter evaluation allows further assessment the importance of the auxiliary data, as it is expected that the fusion models will have less performance degradation, since they can collect information from the auxiliary modality.

\subsection{Retinal Blood Vessel Segmentation}

\subsubsection{Data}
For the RBV segmentation experiments, we used an in-house dataset containing 33 OCT volumes and SLO images from 33 patients with AMD.
All images were captured using a Spectralis OCT device (Heidelberg Engineering, DE).
The 3D OCT scans have an approximate volume of $6 \times 6 \times 1.92$ mm\textsuperscript{3} (en-face height $\times$ en-face width $\times$ depth) and a size of $97 \times 1024 \times 496$ voxels.
The acquisition protocol and the preprocessing steps for the SLO images are the same as for the GA dataset (see Subsection~\ref{sec:ga:data}).
The resulting SLO images have a size of $97 \times 1024$ pixels (en-face).
RBV were annotated on the OCT en-face projections by retinal experts.

\subsubsection{Super-Resolution Setting}

\begin{figure}[t]
\centering
\includegraphics[height=0.15\textwidth]{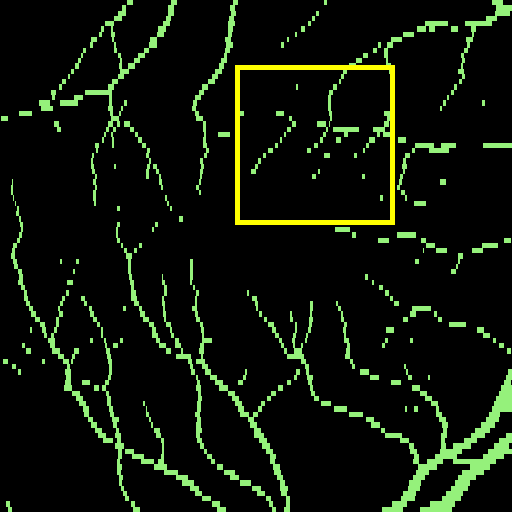}
\includegraphics[height=0.15\textwidth]{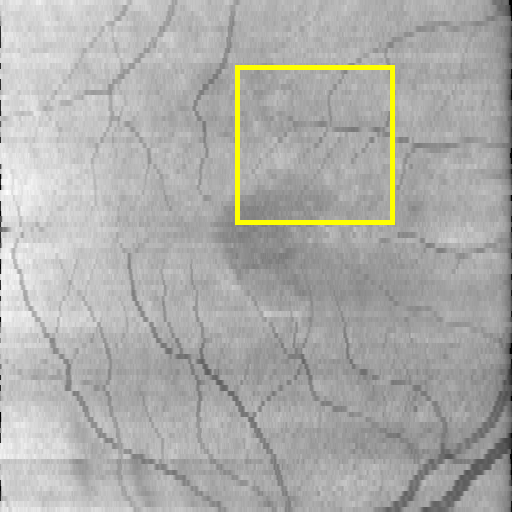}
\includegraphics[height=0.15\textwidth]{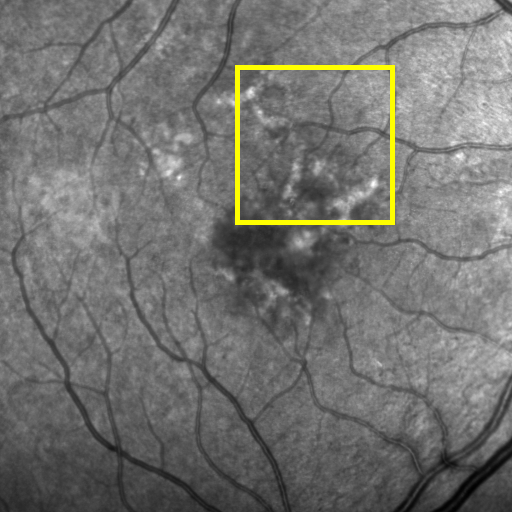} \\
    \vspace{0.1cm}
\includegraphics[height=0.15\textwidth]{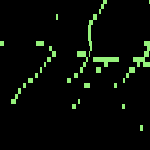}
\includegraphics[height=0.15\textwidth]{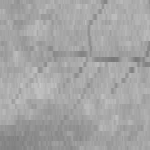}
\includegraphics[height=0.15\textwidth]{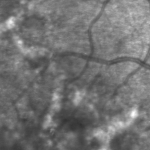}
\caption{%
Example of missing segments in the OCT reference annotations from RBV-S.
From left to right: Reference annotations in green, OCT en-face projection, SLO.
    Zoomed in regions (in yellow, top row) are shown in the bottom row.
}%
\label{fig:super_resolution_rbvs}
\end{figure}

The evaluation of the proposed approach for RBV segmentation presents an important challenge.
RBV annotations were directly done on the OCT.
However, OCT volumes have a low en-face vertical resolution, so some vessel segments are not visible in the OCT and therefore were not annotated.
This leads to a large number of discontinuities in the reference annotation.
In SLO, however, these segments are clearly visible (see Fig.~\ref{fig:super_resolution_rbvs}).
Thus, a method fusing OCT and SLO is expected to segment these segments, which, although they are not annotated in the reference, they are arguably correct when identified as ``vessel'' in a multimodal setting.
To mitigate this problem, we evaluated the proposed approach in a super-resolution setting.
In particular, we trained the models in RBV-S using only half of the slices in the OCT volumes (i.e., 48 slices)
and then evaluated them on the full volumes.
In this way, we could assess whether the fusion models provided higher resolution segmentations, in which more of these missing segments in the OCT were correctly segmented.

\subsubsection{Experiments}

We performed the same experiments and ablation study as for GA segmentation (see Subsection~\ref{sec:ga_segmentation_experiments}), but using only SLO as the second modality, OCT-based annotations, and, due to the reduced dataset size, 20\%, 50\%, and 100\% of the training data.

\subsection{Training and Evaluation Details}\label{sec:training_and_evaluation_details}

As segmentation loss, we used the direct sum of Dice loss~\cite{milletari2016v} and Binary Cross-Entropy (BCE).
These two losses are commonly used for binary segmentation tasks~\cite{milletari2016v,Lachinov_MICCAI_2021,Li_IPN_TMI_2020,Li_IPNv2_arXiv_2020,Orlando_MIA_2020}.
Dice loss is particularly suitable for imbalanced datasets, as it is not affected by class prevalence~\cite{Sudre_Dice_2017}, while BCE is more stable during training and provides a better gradient signal~\cite{yeung2022unified}.
Our code is publicly available on GitHub\footnote{\url{https://github.com/j-morano/multimodal-fusion-fpn}}.

Following previous works~\cite{Lachinov_MICCAI_2021,morano2023selfsupervised}, the OCT volumes were flattened along the Bruch's membrane, and a 3D region of 128 pixels was cropped along the A-scan (vertical axial direction), while the rest of the en-face dimensions remained full size.
Bruch's membrane was localized using the method proposed in~\cite{fazekas2022segmentation}.
In addition, all images were normalized using Z-score normalization.
For data augmentation, we used random flipping and slight multiplicative noise, additive noise, contrast augmentation, and intensity shifts.

\newcolumntype{g}{>{\columncolor{mygray}}l}

\begin{table*}[t!]
    \centering
    \caption{
        GA segmentation results of the different methods on the test set (ours: \textit{Multiscale} and \textit{Late}).
        The table includes both the ablation study (baselines$^\dagger$) and the state-of-the-art comparison.
        The best results are highlighted in \textbf{bold}.
        The p-value resulting from a Wilcoxon signed rank test vs. the FPN/Volume-br (baseline) OCT model is indicated in each case (*: $p < 0.05$, **: $p < 0.01$, ***: $p < 0.001$),
        as well as the difference of the mean with respect to the mean of the best approach (in parentheses).
    }
    \label{tab:ga_segmentation}
    \begin{tabular}{cggggg}
        \toprule
        \rowcolor{white}
        \textbf{Data pct.} & \textbf{Method} & \textbf{Dice} (\%) & $\mathbf{HD_{95}}$ & \textbf{AUROC} (\%) & \textbf{AUPR} (\%) \\
        \midrule

 \multirow{9}{*}{\vspace{-0.1cm}10\%}
    & U-Net FAF
       & $73.98 \pm 22.16$~{\tiny(11.58)} & $0.96 \pm 0.55$~{\tiny(0.36)} & $94.73$~{\tiny(4.12)} & $86.73$~{\tiny(10.14)} \\
    \rowcolor{white}
    & Image-br FAF$^\dagger$
       & $74.51 \pm 22.15$~{\tiny(11.05)} & $1.03 \pm 0.58$~{\tiny(0.43)} & $94.95$~{\tiny(3.90)} & $86.90$~{\tiny(9.98)} \\
    & Image-br SLO$^\dagger$
       & $64.86 \pm 28.92$~{\tiny(20.70)} & $1.36 \pm 0.81$~{\tiny(0.77)} & $88.28$~{\tiny(10.57)} & $69.22$~{\tiny(27.66)} \\
    \rowcolor{white}
    & Lachinov \textit{et al.} OCT
       & $82.35 \pm 17.92$~{\tiny(3.21)} & $0.78 \pm 0.71$~{\tiny(0.18)} & $98.11$~{\tiny(0.75)} & $94.73$~{\tiny(2.14)} \\
    & ReSensNet OCT
       & $82.46 \pm 18.03$~{\tiny(3.10)} & $0.72 \pm 0.71$~{\tiny(0.12)} & $98.17$~{\tiny(0.69)} & $94.91$~{\tiny(1.97)} \\
    \rowcolor{white}
        & FPN/Volume-br OCT$^\dagger$
       & $83.93 \pm 16.20$~{\tiny(1.62)} & $0.68 \pm 0.64$~{\tiny(0.08)} & $98.41$~{\tiny(0.44)} & $95.45$~{\tiny(1.43)} \\
  \cmidrule{2-6}
    & Multiscale OCT+FAF
       & $\mathbf{85.56 \pm 16.09}^{***}$ & $0.60 \pm 0.53^{***}$~{\tiny(0.00)} & $\mathbf{98.85}$ & $\mathbf{96.88}$ \\
    \rowcolor{white}
    & Multiscale OCT+SLO
       & $83.48 \pm 16.79$~{\tiny(2.07)} & $0.67 \pm 0.65$~{\tiny(0.07)} & $98.39$~{\tiny(0.46)} & $95.59$~{\tiny(1.29)} \\
    & Late OCT+FAF
       & $85.20 \pm 16.31^{***}$~{\tiny(0.36)} & $\mathbf{0.60 \pm 0.47}^{***}$ & $98.55$~{\tiny(0.31)} & $95.72$~{\tiny(1.15)} \\
    \rowcolor{white}
    & Late OCT+SLO
       & $84.32 \pm 15.86^{**}$~{\tiny(1.23)} & $0.68 \pm 0.58$~{\tiny(0.08)} & $98.46$~{\tiny(0.40)} & $95.59$~{\tiny(1.28)} \\

  \midrule

 \multirow{9}{*}{\vspace{-0.1cm}20\%}
    & U-Net FAF
       & $76.64 \pm 20.99$~{\tiny(11.12)} & $0.94 \pm 0.58$~{\tiny(0.47)} & $95.61$~{\tiny(3.60)} & $89.39$~{\tiny(8.60)} \\
    \rowcolor{white}
    & Image-br FAF$^\dagger$
       & $74.92 \pm 21.60$~{\tiny(12.84)} & $1.06 \pm 0.64$~{\tiny(0.59)} & $94.86$~{\tiny(4.35)} & $88.02$~{\tiny(9.97)} \\
    & Image-br SLO$^\dagger$
       & $68.30 \pm 25.88$~{\tiny(19.47)} & $1.21 \pm 0.72$~{\tiny(0.74)} & $92.05$~{\tiny(7.16)} & $82.65$~{\tiny(15.34)} \\
    \rowcolor{white}
    & Lachinov \textit{et al.} OCT
       & $85.29 \pm 17.33$~{\tiny(2.48)} & $0.65 \pm 0.82$~{\tiny(0.18)} & $98.90$~{\tiny(0.31)} & $97.23$~{\tiny(0.76)} \\
    & ReSensNet OCT
       & $85.71 \pm 16.20$~{\tiny(2.06)} & $0.58 \pm 0.71$~{\tiny(0.11)} & $98.94$~{\tiny(0.27)} & $97.25$~{\tiny(0.74)} \\
    \rowcolor{white}
    & FPN/Volume-br OCT$^\dagger$
       & $87.13 \pm 13.83$~{\tiny(0.63)} & $0.50 \pm 0.45$~{\tiny(0.03)} & $99.05$~{\tiny(0.16)} & $97.67$~{\tiny(0.32)} \\
  \cmidrule{2-6}
    & Multiscale OCT+FAF
       & $\mathbf{87.77 \pm 14.70}^{***}$ & $\mathbf{0.47 \pm 0.46}^{***}$ & $\mathbf{99.21}$ & $\mathbf{97.99}$ \\
    \rowcolor{white}
    & Multiscale OCT+SLO
       & $86.54 \pm 14.83$~{\tiny(1.23)} & $0.54 \pm 0.49$~{\tiny(0.07)} & $99.02$~{\tiny(0.18)} & $97.32$~{\tiny(0.67)} \\
    & Late OCT+FAF
       & $86.98 \pm 15.37^{***}$~{\tiny(0.78)} & $0.50 \pm 0.46$~{\tiny(0.03)} & $99.09$~{\tiny(0.12)} & $97.50$~{\tiny(0.49)} \\
    \rowcolor{white}
    & Late OCT+SLO
       & $86.41 \pm 15.53$~{\tiny(1.36)} & $0.50 \pm 0.46$~{\tiny(0.03)} & $99.01$~{\tiny(0.20)} & $97.42$~{\tiny(0.57)} \\

  \midrule

 \multirow{9}{*}{\vspace{-0.1cm}100\%}
    & U-Net FAF
       & $80.93 \pm 18.98$~{\tiny(9.26)} & $0.80 \pm 0.58$~{\tiny(0.42)} & $97.26$~{\tiny(2.34)} & $93.08$~{\tiny(5.87)} \\
    \rowcolor{white}
    & Image-br FAF$^\dagger$
       & $78.06 \pm 20.38$~{\tiny(12.13)} & $0.96 \pm 0.61$~{\tiny(0.58)} & $96.57$~{\tiny(3.03)} & $91.90$~{\tiny(7.05)} \\
    & Image-br SLO$^\dagger$
       & $70.03 \pm 26.79$~{\tiny(20.16)} & $1.17 \pm 0.76$~{\tiny(0.79)} & $93.13$~{\tiny(6.47)} & $84.96$~{\tiny(13.99)} \\
    \rowcolor{white}
    & Lachinov \textit{et al.} OCT
       & $88.61 \pm 11.39$~{\tiny(1.58)} & $0.48 \pm 0.45$~{\tiny(0.10)} & $99.21$~{\tiny(0.39)} & $98.03$~{\tiny(0.91)} \\
    & ReSensNet OCT
       & $88.57 \pm 13.21$~{\tiny(1.62)} & $0.42 \pm 0.46$~{\tiny(0.04)} & $99.33$~{\tiny(0.27)} & $98.23$~{\tiny(0.72)} \\
    \rowcolor{white}
    & FPN/Volume-br OCT$^\dagger$
       & $89.13 \pm 11.93$~{\tiny(1.06)} & $0.40 \pm 0.38$~{\tiny(0.01)} & $99.39$~{\tiny(0.21)} & $98.37$~{\tiny(0.58)} \\
  \cmidrule{2-6}
    & Multiscale OCT+FAF
       & $\mathbf{90.19 \pm 12.21}^{***}$ & $\mathbf{0.38 \pm 0.42}$ & $\mathbf{99.60}$ & $\mathbf{98.95}$ \\
    \rowcolor{white}
    & Multiscale OCT+SLO
       & $89.15 \pm 12.26^{***}$~{\tiny(1.04)} & $0.45 \pm 0.57$~{\tiny(0.06)} & $99.39$~{\tiny(0.21)} & $98.40$~{\tiny(0.55)} \\
    & Late OCT+FAF
       & $89.91 \pm 11.62^{***}$~{\tiny(0.28)} & $0.40 \pm 0.42$~{\tiny(0.02)} & $99.48$~{\tiny(0.12)} & $98.59$~{\tiny(0.35)} \\
    \rowcolor{white}
    & Late OCT+SLO
       & $89.08 \pm 11.39$~{\tiny(1.11)} & $0.42 \pm 0.37$~{\tiny(0.03)} & $99.33$~{\tiny(0.27)} & $98.22$~{\tiny(0.73)} \\

    \bottomrule\\

    \toprule
    \multicolumn{6}{c}{\textbf{OCT-based annotations subset}} \\
    \midrule

  \rowcolor{white}
  \multirow{4}{*}{10\%}
    & FPN/Volume-br OCT$^\dagger$
       & $81.10 \pm 18.54$~{\tiny(1.26)} & $0.64 \pm 0.52$~{\tiny(0.01)} & $98.55$~{\tiny(0.35)} & $95.05$~{\tiny(1.39)} \\
  \cmidrule{2-6}
    & Multiscale OCT+FAF
       & $\mathbf{82.36 \pm 18.18}^{**}$ & $0.74 \pm 0.65$~{\tiny(0.10)} & $\mathbf{98.90}$ & $\mathbf{96.44}$ \\
    \rowcolor{white}
    & Multiscale OCT+SLO
       & $80.64 \pm 20.15$~{\tiny(1.73)} & $\mathbf{0.64 \pm 0.69}$ & $98.40$~{\tiny(0.50)} & $94.81$~{\tiny(1.63)} \\
    & Late OCT+FAF
       & $81.99 \pm 18.23$~{\tiny(0.37)} & $0.77 \pm 0.68$~{\tiny(0.13)} & $98.50$~{\tiny(0.40)} & $94.77$~{\tiny(1.67)} \\
    \rowcolor{white}
    & Late OCT+SLO
       & $81.49 \pm 18.85$~{\tiny(0.88)} & $0.71 \pm 0.78$~{\tiny(0.07)} & $98.51$~{\tiny(0.39)} & $94.95$~{\tiny(1.49)} \\

  \midrule

  \rowcolor{white}
  \multirow{4}{*}{20\%}
    & FPN/Volume-br OCT$^\dagger$
       & $84.55 \pm 16.18$~{\tiny(0.98)} & $0.49 \pm 0.43$~{\tiny(0.03)} & $99.14$~{\tiny(0.21)} & $97.61$~{\tiny(0.48)} \\
  \cmidrule{2-6}
    & Multiscale OCT+FAF
       & $\mathbf{85.53 \pm 16.08}^{**}$ & $\mathbf{0.45 \pm 0.40}$ & $\mathbf{99.34}$ & $\mathbf{98.09}$ \\
    \rowcolor{white}
    & Multiscale OCT+SLO
       & $83.94 \pm 17.23$~{\tiny(1.59)} & $0.51 \pm 0.42$~{\tiny(0.05)} & $99.04$~{\tiny(0.31)} & $97.15$~{\tiny(0.94)} \\
    & Late OCT+FAF
       & $84.14 \pm 17.82$~{\tiny(1.39)} & $0.52 \pm 0.43$~{\tiny(0.07)} & $99.17$~{\tiny(0.18)} & $97.28$~{\tiny(0.81)} \\
    \rowcolor{white}
    & Late OCT+SLO
       & $82.66 \pm 21.17$~{\tiny(2.87)} & $0.50 \pm 0.44$~{\tiny(0.04)} & $99.06$~{\tiny(0.28)} & $97.26$~{\tiny(0.84)} \\

  \midrule

  \rowcolor{white}
  \multirow{4}{*}{100\%}
    & FPN/Volume-br OCT$^\dagger$
       & $87.29 \pm 12.62$~{\tiny(1.61)} & $0.38 \pm 0.30$~{\tiny(0.02)} & $99.41$~{\tiny(0.20)} & $98.14$~{\tiny(0.64)} \\
  \cmidrule{2-6}
    & Multiscale OCT+FAF
       & $88.45 \pm 11.92^{**}$~{\tiny(0.45)} & $\mathbf{0.36 \pm 0.36}$ & $\mathbf{99.61}$ & $\mathbf{98.77}$ \\
    \rowcolor{white}
    & Multiscale OCT+SLO
       & $86.83 \pm 12.75$~{\tiny(2.08)} & $0.55 \pm 0.74$~{\tiny(0.19)} & $99.37$~{\tiny(0.24)} & $98.03$~{\tiny(0.75)} \\
    & Late OCT+FAF
       & $\mathbf{88.90 \pm 9.44}^{**}$ & $0.37 \pm 0.26$~{\tiny(0.01)} & $99.52$~{\tiny(0.09)} & $98.40$~{\tiny(0.37)} \\
    \rowcolor{white}
    & Late OCT+SLO
       & $88.03 \pm 10.94$~{\tiny(0.87)} & $0.38 \pm 0.31$~{\tiny(0.02)} & $99.36$~{\tiny(0.25)} & $97.96$~{\tiny(0.82)} \\

  \bottomrule

  \end{tabular}
\end{table*}

Models were trained for 800 epochs using SGD with a learning rate of $0.1$ and a momentum of $0.9$.
Batch size was set to 8.
All the experiments were performed on a server
with two AMD EPYC 7443 24-Core CPUs, 1024GB of RAM, and four NVIDIA RTX A6000s, of which only one was used.

All datasets were split patient-wise into training (60\%), validation (10\%) and test (30\%) sets.
The experiments with limited training data were performed by selecting a subset of patients from the training set, so that the smaller sets were always subsets of the larger ones.
The validation and test sets were kept fixed for all experiments.

For the experiments with noisy input data, we used different levels of random masking (cutout) with rectangular masks of size $0.95H \times 0.1W \times 0.1D$, where $H$, $W$, and $D$ are the en-face height, en-face width, and depth of the volume.
The values of the masked voxels were randomly sampled from a uniform distribution in the range $[\mu-0.1\mu, \mu+0.1\mu]$, where $\mu$ is the mean of the voxel intensities.

To reduce inference variability, we averaged the predictions of five models corresponding to the five checkpoints with the lowest Dice error in the validation set.

Segmentation performance was evaluated as in previous works~\cite{Lachinov_MICCAI_2021,morano2023selfsupervised,morano2021simultaneous,chen2021retinal}
by comparing the predicted segmentation masks with the reference
via
Dice score, area under receiver operating characteristic curve (AUROC), and area under precision-recall curve (AUPR).
Moreover, GA segmentations were evaluated via the 95\textsuperscript{th} percentile of the Hausdorff distance (\HD) between predicted and manual masks.
To assess statistical significance, the Wilcoxon signed-rank test was used.

\section{Results and Discussion}%
\label{sec:results_and_discussion}

\subsection{Geographic Atrophy Segmentation}%
\label{sec:results_ga_segmentation}

Table~\ref{tab:ga_segmentation} shows the average Dice score and \HD \ values as well as the AUROC and AUPR values of the different models on the GA test set.
The proposed Multiscale method using OCT and FAF images (Multiscale OCT+FAF) was always the best performing method across all training data pcts., state-of-the-art methods and fusion schemes.
The differences were particularly pronounced in scenarios with label scarcity.
For example, using 10\% of the training data, Multiscale OCT+FAF significantly surpassed the baseline FPN/Volume-br OCT in terms of Dice ($+1.58\%$), \HD \ ($-0.08$), AUROC ($+0.44\%$) and AUPR ($+1.43\%$).
With respect to Lachinov \textit{et al.}, the differences were even more pronounced: $+2.21\%$ Dice, $-0.18$ \HD, $+0.74\%$ AUROC, and $+2.15\%$ AUPR.
Moreover, to achieve results similar to those of Lachinov \textit{et al.} with 100\% of the training data, Multiscale OCT+FAF required only 20\%.

It is further observed that the proposed Late Fusion method using OCT and FAF images (Late OCT+FAF) also significantly outperformed the state of the art and, in most cases, FPN/Volume-br OCT.
However, the differences were smaller than those obtained for the Multiscale method.
Lastly, the proposed method using OCT and SLO outperformed the baseline only in a few cases, and the differences were not statistically significant, indicating that the SLO modality is not as useful as FAF for the task of segmenting GA lesions.

Regarding the models trained with FAF or SLO alone, the performance of Image-br is similar to that of U-Net, but far below that of FPN/Volume-br, reinforcing the idea that OCT is the most appropriate modality for this task~\cite{mai2022comparison}.

\begin{figure}
    \includegraphics[width=0.95\linewidth]{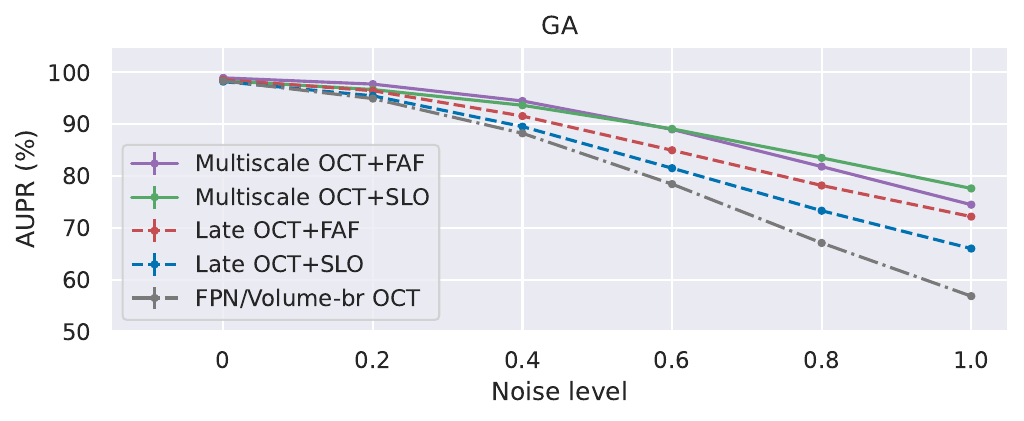}
    \includegraphics[width=0.95\linewidth]{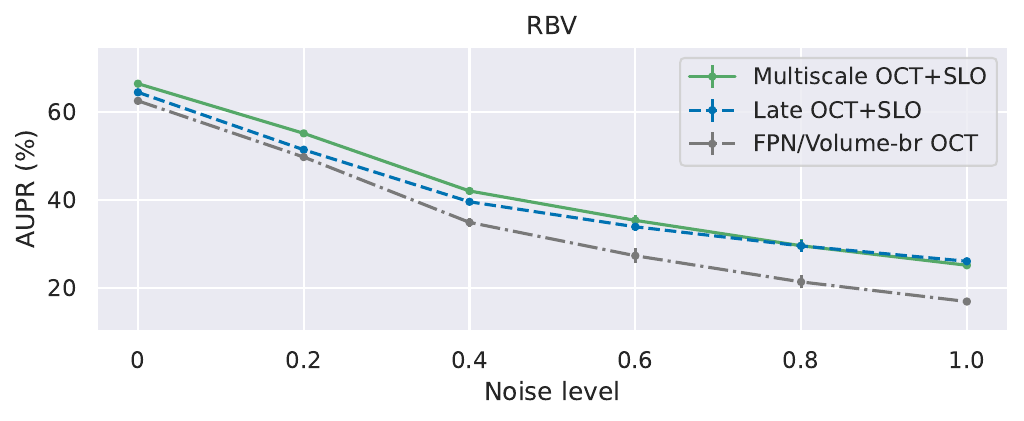}
    \caption{%
        AUPR values obtained by fusion and FPN/Volume-br methods trained on 100\% of the training data and evaluated using different levels and types of noise on the OCT.
    }
    \label{fig:noise}
\end{figure}

Similar results were obtained when the models were evaluated on the OCT-based annotations (see Table~\ref{tab:ga_segmentation}, bottom), confirming that the improvement of the proposed methods does not depend on the specific annotation modality.
However, in this case, Late OCT+FAF was found to be significantly better than FPN/Volume-br OCT only when using 100\% of the training data, confirming the intuition that using deep features at different scales is better for exploiting the complementary information provided by the second modality.
Furthermore, the evaluation of the models under noise on the OCT (see Fig.~\ref{fig:noise}, top) shows that the proposed methods, and especially the Multiscale methods, are much more robust than the monomodal baseline method FPN/Volume-br OCT.

\begin{figure*}[!ht]
    \setlength{\tabcolsep}{5pt}
    \centering
    \begin{tabular}{cccccc}
        & & {\scriptsize 92.07 / 1.45} & {\scriptsize 88.62 / 1.45} & {\scriptsize \textbf{98.34 / 0.12}} & {\scriptsize 94.36 / 0.28} \\

        \includegraphics[width=0.14\linewidth]{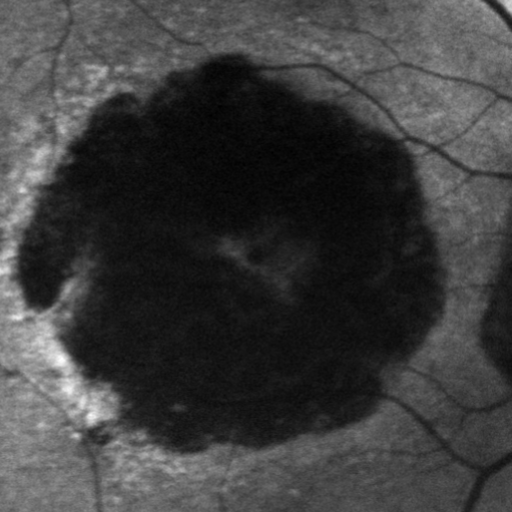}
        & \includegraphics[width=0.14\linewidth]{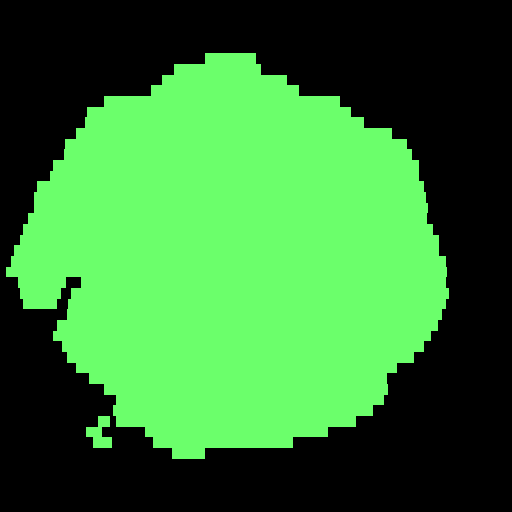}
        & \includegraphics[width=0.14\linewidth]{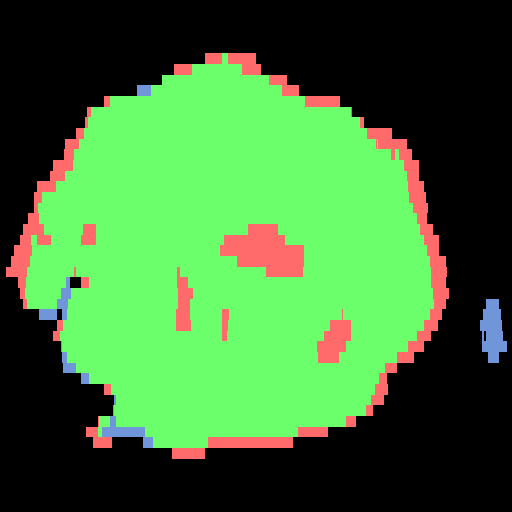}
        & \includegraphics[width=0.14\linewidth]{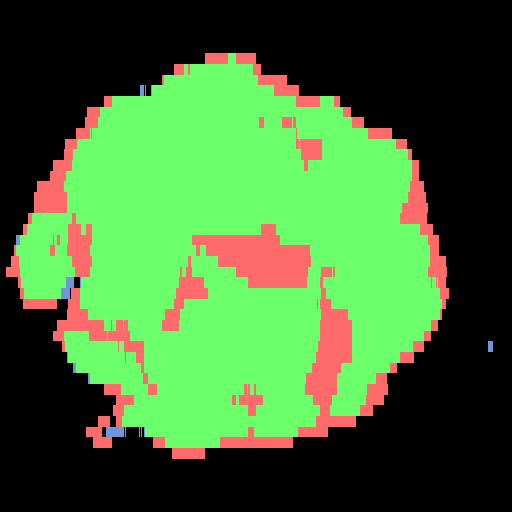}
        & \includegraphics[width=0.14\linewidth]{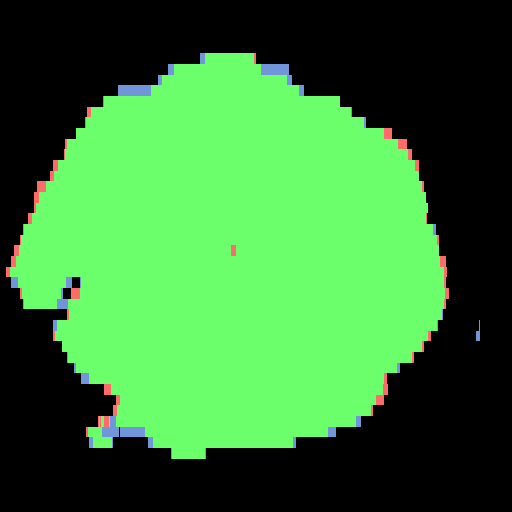}
        & \includegraphics[width=0.14\linewidth]{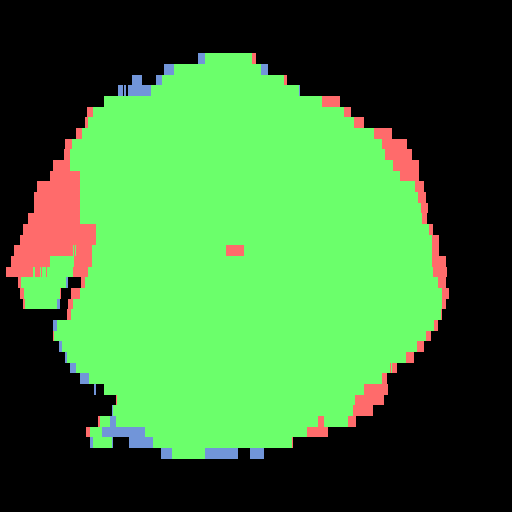} \\

        & & {\scriptsize 65.58 / 1.09} & {\scriptsize 70.55 / 0.83} & {\scriptsize \textbf{81.0 / 0.17}} & {\scriptsize 79.74 / 0.62} \\

        \includegraphics[width=0.14\linewidth]{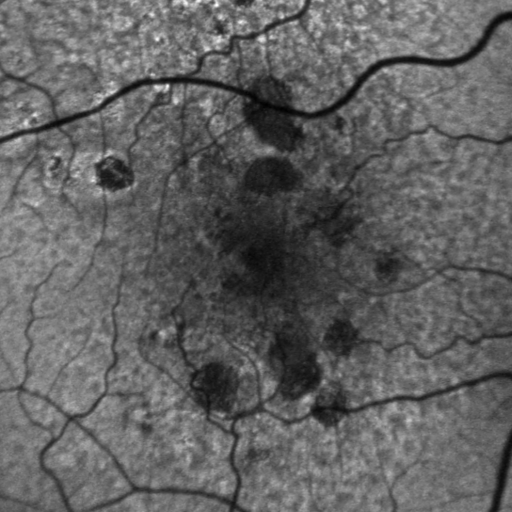}
        & \includegraphics[width=0.14\linewidth]{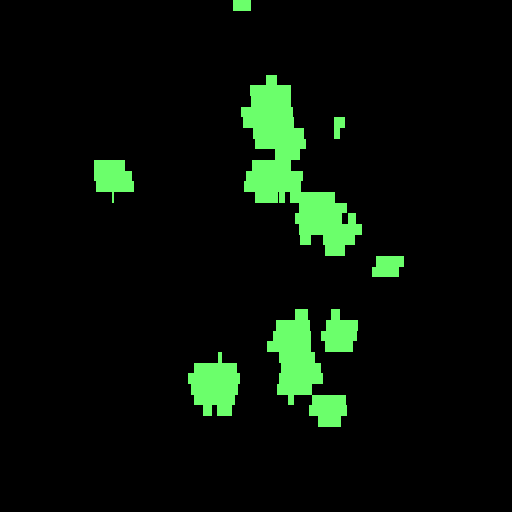}
        & \includegraphics[width=0.14\linewidth]{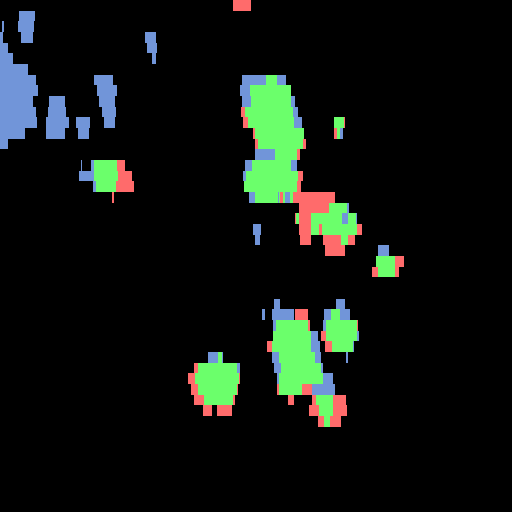}
        & \includegraphics[width=0.14\linewidth]{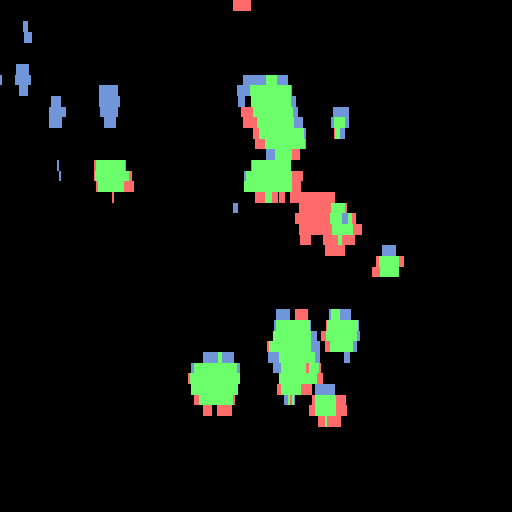}
        & \includegraphics[width=0.14\linewidth]{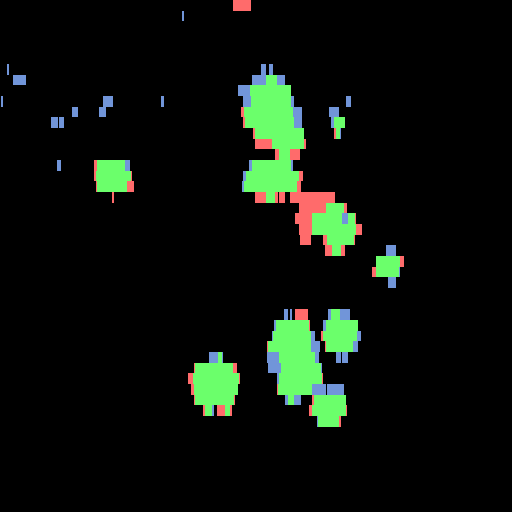}
        & \includegraphics[width=0.14\linewidth]{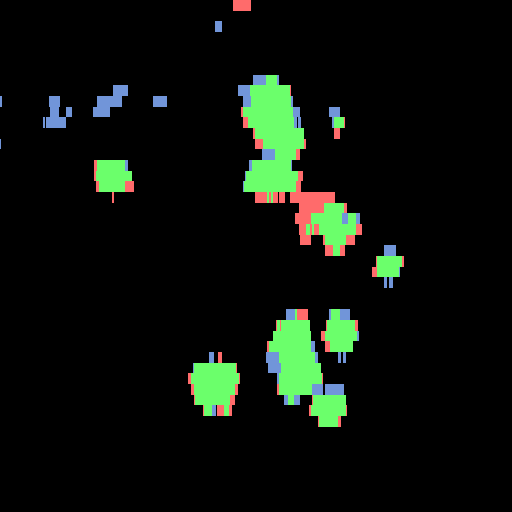} \\

        FAF
        & Reference
        & Lachinov \textit{et al.}
        & FPN/Volume-br
        & Multiscale
        & Late \\

        & & & OCT & OCT+FAF & OCT+FAF \\
    \end{tabular}
    \caption{
        Qualitative results obtained by the different methods on the test set using the 10\% of the training data.
        \textcolor{OliveGreen}{\textbf{True positives}} are depicted in green; \textbf{true negatives}, in black; \textcolor{Blue}{\textbf{false positives}}, in blue; and \textcolor{BrickRed}{\textbf{false negatives}}, in red.
        Dice~/~\HD \ values are shown above each prediction.
    }
    \label{fig:qualitative_ga}
\end{figure*}

Fig.~\ref{fig:qualitative_ga} shows qualitative results obtained by some of the methods listed in Table~\ref{tab:ga_segmentation} using 10\% of the training data.
These results are in line with the quantitative results shown in Table~\ref{tab:ga_segmentation}.
The proposed Multiscale OCT+FAF method was able to much better delineate the GA lesions than FPN/Volume-br OCT and Lachinov \textit{et al.}~\cite{Lachinov_MICCAI_2021}.
It is particularly meaningful the difference regarding the false negatives (in \textcolor{BrickRed}{red}). %
The same is observed for the Late OCT+FAF approach, although the differences are smaller, which is in concordance with the quantitative results.

The results of the GA segmentation experiments show that fusing OCT and FAF images with the proposed framework is beneficial regardless of the fusion approach.
Moreover, the greater robustness of the fusion models to noisy OCTs suggests that fusion approaches effectively exploit the complementary information provided by the second modality.
The improvement is even greater when the amount of labeled data is scarce.
This is in contrast to the fusion of OCT and SLO, where the improvement was neither consistent nor significant.
This is probably due to the fact that SLO is not as informative as FAF for the assessment of GA, confirming that an adequate choice of the modality to be fused is crucial for the success of the fusion approach.
Finally, the results also show that the proposed Multiscale Fusion approach was considerably more effective than the Late Fusion approach, suggesting that the fusion in the intermediate feature representation allows for a more meaningful use of the available feature spaces.
This observation is consistent with the results of Dolz \textit{et al.} in \cite{Dolz_CSI_2019}.

\subsection{Retinal Blood Vessel Segmentation}%
\label{sec:rbv_segmentation}

The results of the different methods on the RBV test set in terms of Dice, AUROC, and AUPR are shown in Table~\ref{tab:rbv_segmentation}.
\begin{table}[tbp]
    \centering
    \caption{
        RBV segmentation results of the different methods on the test set.
        See the legend in the caption of Table~\ref{tab:ga_segmentation}.
    }
    \label{tab:rbv_segmentation}
    \resizebox{\linewidth}{!}{%
    \begin{tabular}{cgggg}
        \toprule
        \rowcolor{white}
        \textbf{Data pct.} & \textbf{Method} & \textbf{Dice} (\%) & \textbf{AUROC} (\%) & \textbf{AUPR} (\%) \\
        \midrule

  \rowcolor{white}
 \multirow{7}{*}{20\%}
    & U-Net SLO
       & $39.50 \pm 5.33$~{\tiny(20.23)} & $85.14$~{\tiny(7.32)} & $32.54$~{\tiny(32.33)} \\
    & Image-br SLO$^\dagger$
       & $36.08 \pm 8.05$~{\tiny(23.65)} & $82.83$~{\tiny(9.64)} & $30.85$~{\tiny(34.01)} \\
    \rowcolor{white}
    & Lachinov \textit{et al.} OCT
       & $53.34 \pm 12.73$~{\tiny(6.39)} & $90.42$~{\tiny(2.04)} & $55.75$~{\tiny(9.11)} \\
    & ReSensNet OCT
       & $55.10 \pm 12.40$~{\tiny(4.64)} & $89.21$~{\tiny(3.25)} & $59.56$~{\tiny(5.30)} \\
    \rowcolor{white}
    & FPN/Volume-br OCT$^\dagger$
       & $56.63 \pm 13.62$~{\tiny(3.11)} & $91.75$~{\tiny(0.71)} & $60.93$~{\tiny(3.93)} \\
  \cmidrule{2-5}
    & Multiscale OCT+SLO
       & $\mathbf{59.74 \pm 6.97}$ & $\mathbf{92.46}$ & $\mathbf{64.86}$ \\
    \rowcolor{white}
    & Late OCT+SLO
       & $56.61 \pm 9.29$~{\tiny(3.12)} & $90.91$~{\tiny(1.55)} & $61.77$~{\tiny(3.09)} \\

  \midrule

  \rowcolor{white}
 \multirow{7}{*}{50\%}
    & U-Net SLO
       & $40.26 \pm 5.73$~{\tiny(26.70)} & $85.01$~{\tiny(9.95)} & $33.84$~{\tiny(38.55)} \\
    & Image-br SLO$^\dagger$
       & $39.95 \pm 8.01$~{\tiny(27.01)} & $85.23$~{\tiny(9.73)} & $35.08$~{\tiny(37.31)} \\
  \rowcolor{white}
    & Lachinov \textit{et al.} OCT
       & $61.03 \pm 10.14$~{\tiny(5.94)} & $93.29$~{\tiny(1.67)} & $66.16$~{\tiny(6.22)} \\
    & ReSensNet OCT
       & $63.09 \pm 9.12$~{\tiny(3.88)} & $92.85$~{\tiny(2.11)} & $66.48$~{\tiny(5.90)} \\
  \rowcolor{white}
    & FPN/Volume-br OCT$^\dagger$
       & $65.73 \pm 8.84$~{\tiny(1.23)} & $93.97$~{\tiny(1.00)} & $70.53$~{\tiny(1.86)} \\
  \cmidrule{2-5}
    & Multiscale OCT+SLO
       & $\mathbf{66.97 \pm 6.15}$ & $\mathbf{94.96}$ & $\mathbf{72.39}$ \\
  \rowcolor{white}
    & Late OCT+SLO
       & $65.59 \pm 7.09$~{\tiny(1.38)} & $94.08$~{\tiny(0.89)} & $70.91$~{\tiny(1.47)} \\

  \midrule

    \rowcolor{white}
 \multirow{7}{*}{100\%}
    & U-Net SLO
       & $38.07 \pm 7.46$~{\tiny(28.45)} & $85.64$~{\tiny(9.66)} & $34.82$~{\tiny(38.33)} \\
    & Image-br SLO$^\dagger$
       & $40.38 \pm 8.97$~{\tiny(26.14)} & $86.44$~{\tiny(8.86)} & $36.93$~{\tiny(36.22)} \\
    \rowcolor{white}
    & Lachinov \textit{et al.} OCT
       & $64.27 \pm 8.73$~{\tiny(2.25)} & $95.06$~{\tiny(0.24)} & $70.15$~{\tiny(3.01)} \\
    & ReSensNet OCT
       & $58.79 \pm 11.27$~{\tiny(7.73)} & $93.12$~{\tiny(2.18)} & $65.63$~{\tiny(7.52)} \\
    \rowcolor{white}
    & FPN/Volume-br OCT$^\dagger$
       & $62.60 \pm 8.79$~{\tiny(3.92)} & $95.01$~{\tiny(0.29)} & $70.61$~{\tiny(2.54)} \\
  \cmidrule{2-5}
    & Multiscale OCT+SLO
       & $\mathbf{66.52 \pm 8.30}^{**}$ & $\mathbf{95.30}$ & $\mathbf{73.15}$ \\
    \rowcolor{white}
    & Late OCT+SLO
       & $64.55 \pm 7.14$~{\tiny(1.96)} & $95.28$~{\tiny(0.02)} & $70.35$~{\tiny(2.80)} \\

        \bottomrule
    \end{tabular}
    }
\end{table}
\begin{figure*}[tbhp]
    \setlength{\tabcolsep}{5pt}
    \centering
    \begin{tabular}{cccccc}
        & & {\scriptsize 32.18} & {\scriptsize 33.93} & {\scriptsize \textbf{54.98}} & {\scriptsize 44.54} \\

        \includegraphics[width=0.14\linewidth]{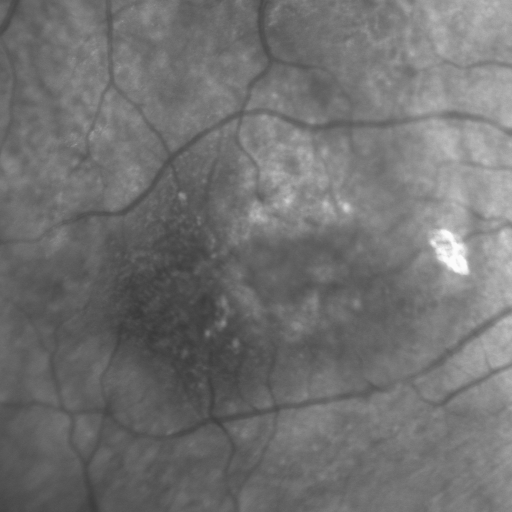}
        & \includegraphics[width=0.14\linewidth]{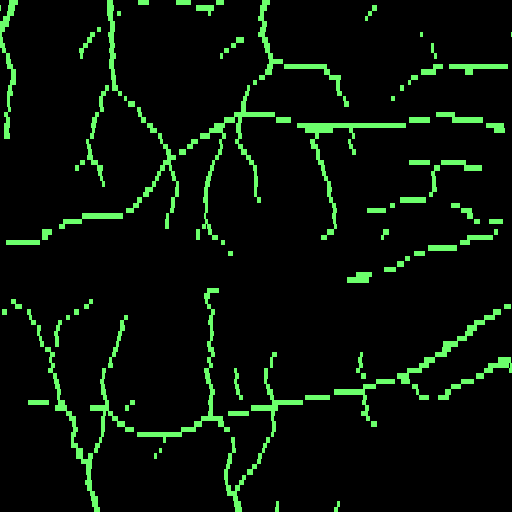}
        & \includegraphics[width=0.14\linewidth]{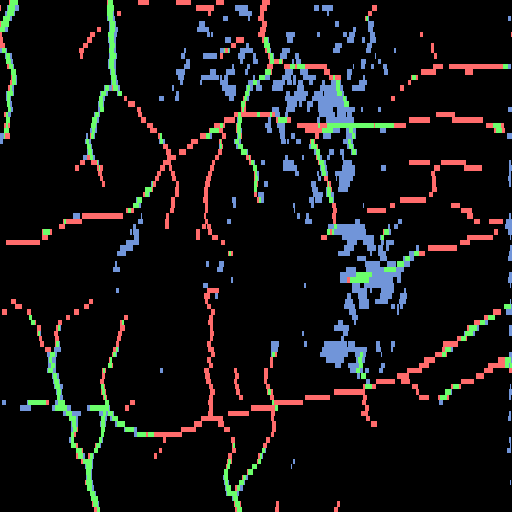}
        & \includegraphics[width=0.14\linewidth]{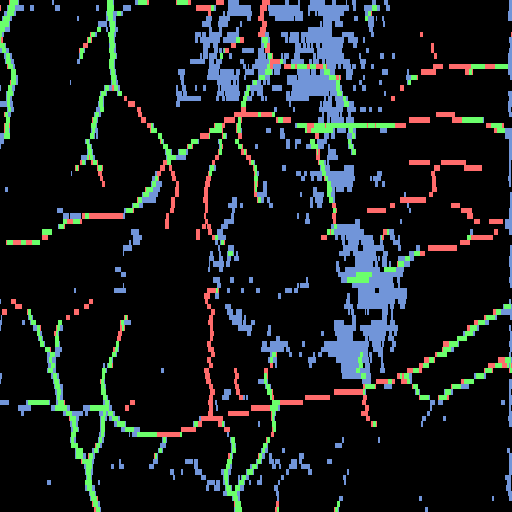}
        & \includegraphics[width=0.14\linewidth]{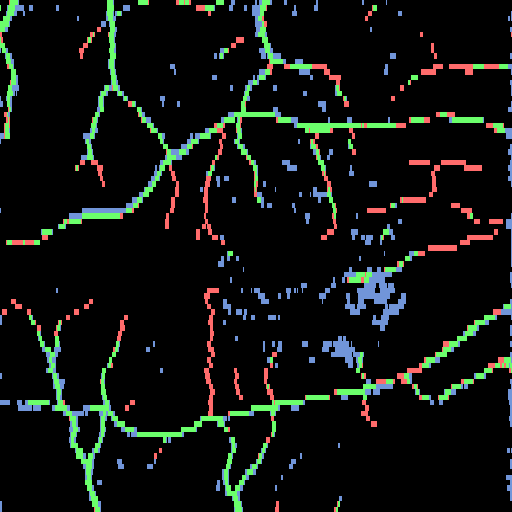}
        & \includegraphics[width=0.14\linewidth]{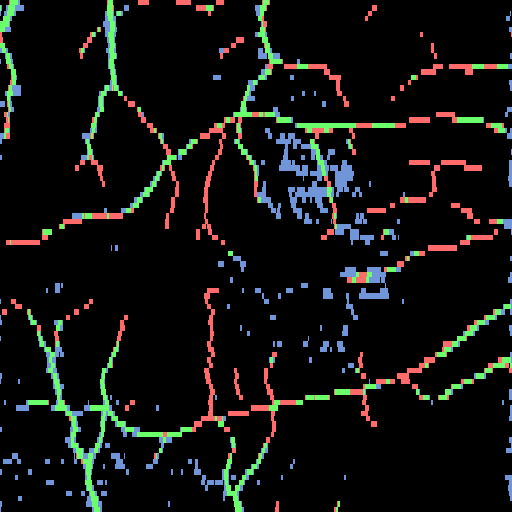} \\

        & & {\scriptsize 58.29} & {\scriptsize 66.05} & {\scriptsize \textbf{66.3}} & {\scriptsize 64.95} \\

        \includegraphics[width=0.14\linewidth]{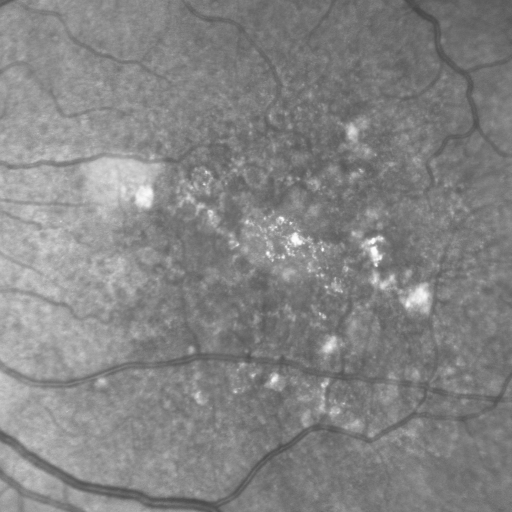}
        & \includegraphics[width=0.14\linewidth]{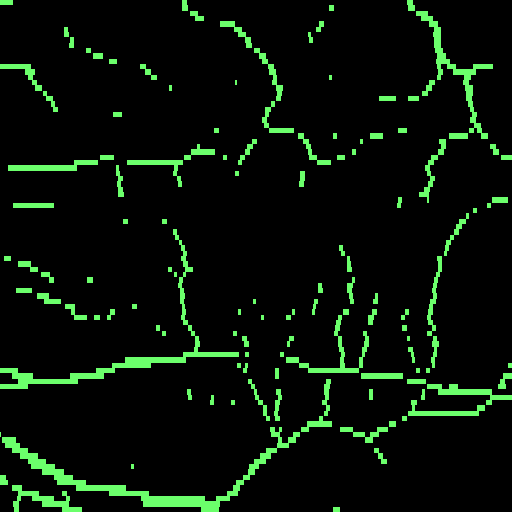}
        & \includegraphics[width=0.14\linewidth]{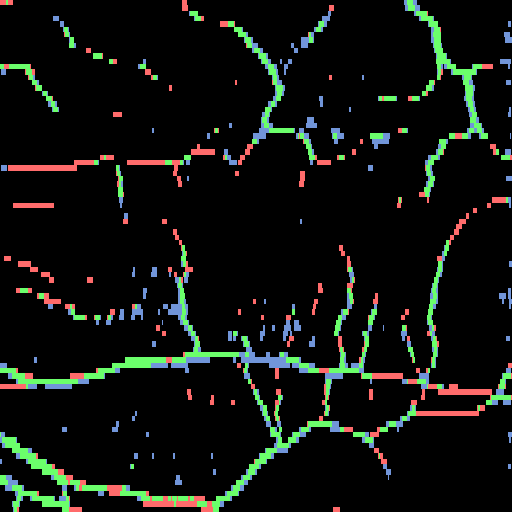}
        & \includegraphics[width=0.14\linewidth]{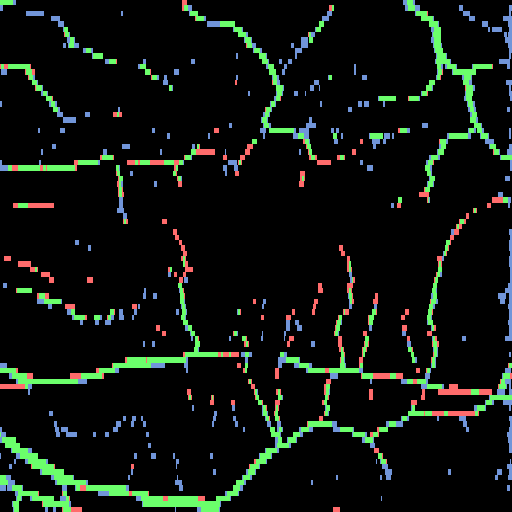}
        & \includegraphics[width=0.14\linewidth]{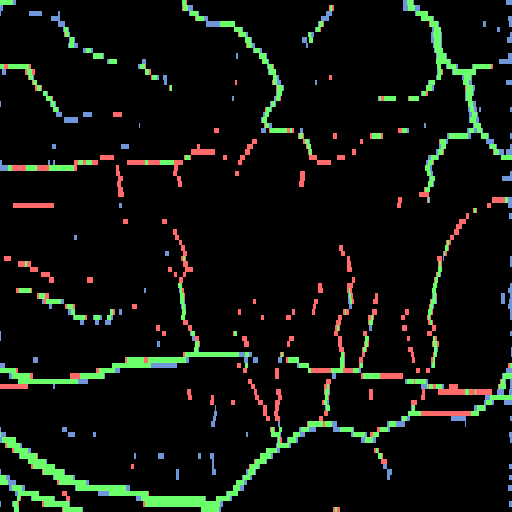}
        & \includegraphics[width=0.14\linewidth]{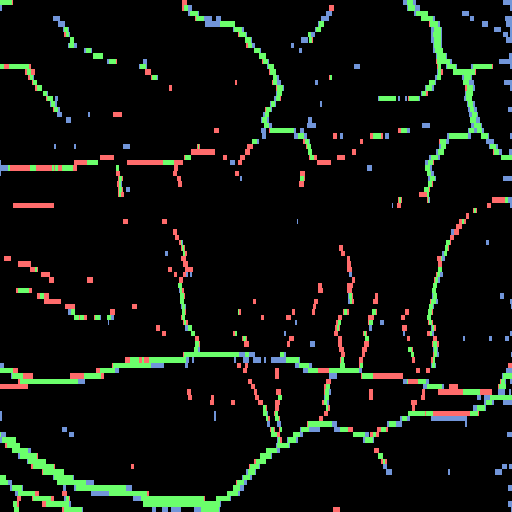} \\

        SLO
        & Reference
        & Lachinov \textit{et al.}
        & FPN/Volume-br
        & Multiscale
        & Late \\

        & & & OCT & OCT+SLO & OCT+SLO \\
    \end{tabular}
    \caption{
        Qualitative results obtained by different methods on the test set using 20\% of the training data.
        \textcolor{OliveGreen}{\textbf{True positives}} are depicted in green; \textbf{true negatives}, in black; \textcolor{Blue}{\textbf{false positives}}, in blue; and \textcolor{BrickRed}{\textbf{false negatives}}, in red.
        Dice values are shown above each prediction.
    }
    \label{fig:qualitative_rbv}
\end{figure*}
The proposed Multiscale Fusion method outperformed baseline and state-of-the-art methods in all cases.
For example, using 20\% of the training data, Multiscale Fusion improved FPN/Volume-br by $3.11\%$ in terms of Dice and by $3.93\%$ in terms of AUPR.
Using 100\%, the improvement is $3.92\%$ in terms of Dice and $2.54\%$ in terms of AUPR.
In this case, the difference was found statistically significant.
With respect to Lachinov \textit{et al.}, the differences were even higher.
Using 20\% of the training data, the proposed Multiscale Fusion method improved the Dice score by $6.40\%$ and the AUPR by $9.11\%$.

As in the case of GA segmentation, the evaluation of the models under noise on the OCT (see Fig.~\ref{fig:noise}, bottom) shows that the proposed methods are more robust to noise than the FPN/Volume-br method, suggesting that they effectively exploit the complementary information from the extra modality.

These results demonstrate the effectiveness of the proposed framework for improving the segmentation of RBV by fusing OCT and SLO images.
This is particularly interesting since the SLO modality comes at no additional cost, as it is already acquired along with the OCT (and automatically co-registered with it by the device) in most clinical settings.
In the experiments, however, the improvement was not as significant as in the case of GA segmentation, as only the Multiscale Fusion approach consistently outperformed the baseline, and, in most cases, by non-significant differences.
This can be explained by the fact that the Late Fusion approach relies heavily on the individual feature extraction capabilities of the Image branch, and SLO images \emph{alone} are not particularly discriminative for the task of RBV segmentation, as suggested by the poor performance of U-Net SLO and Image-br SLO.
This, in the end, caused the Late Fusion models to use lower quality features (from the Image branch) for the final segmentation, affecting the performance of the method.

\begin{figure}[tbhp]
    \centering
    \begin{tabularx}{\linewidth}{cX}
        (a) & \multicolumn{1}{m{0.92\linewidth}}{\includegraphics[width=0.92\linewidth]{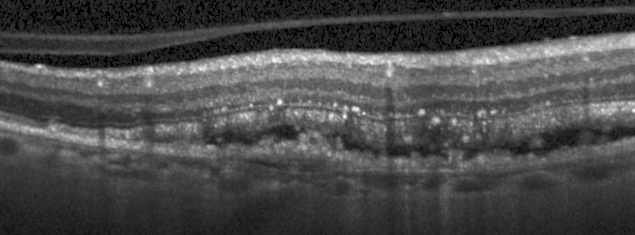}} \\
        (b) & \multicolumn{1}{m{0.92\linewidth}}{\includegraphics[width=0.92\linewidth]{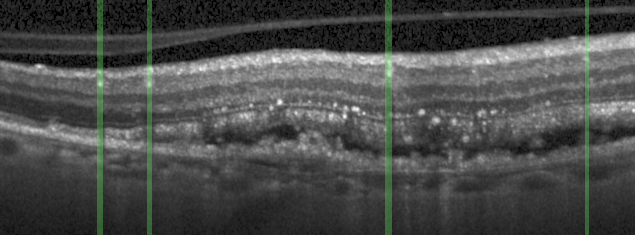}} \\
        (c) & \multicolumn{1}{m{0.92\linewidth}}{\includegraphics[width=0.92\linewidth]{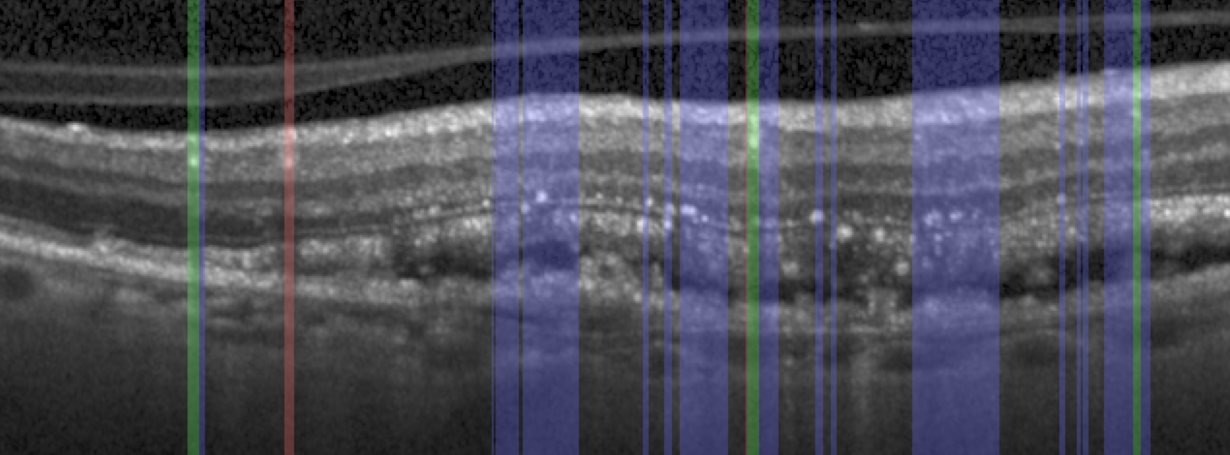}} \\
        (d) & \multicolumn{1}{m{0.92\linewidth}}{\includegraphics[width=0.92\linewidth]{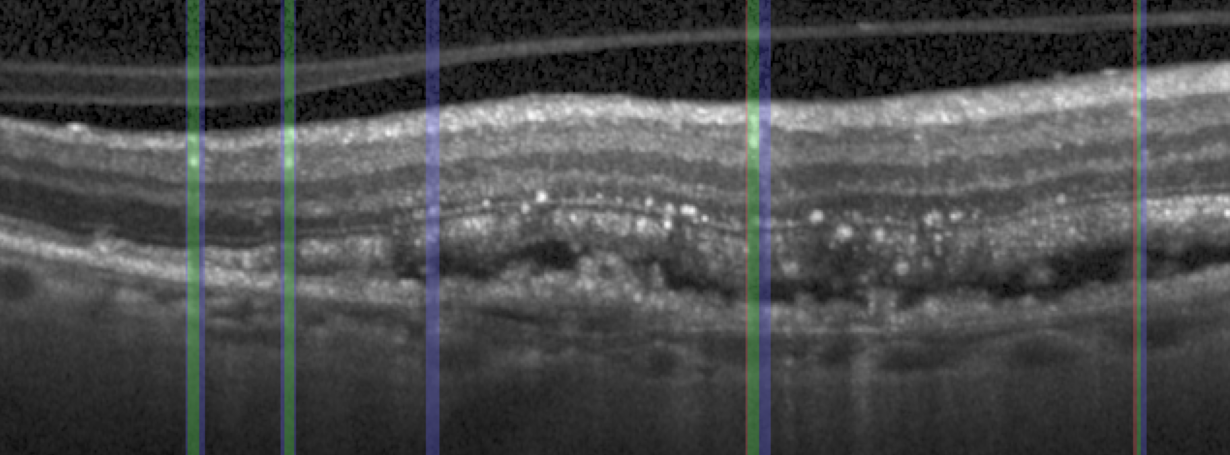}} \\
    \end{tabularx}
    \caption{
        Qualitative results on OCT slices.
        OCT slice \#12 of the sample shown in the first row of Fig.~\ref{fig:qualitative_rbv} with the reference and the predictions of the different models overlaid.
        (a) OCT slice; (b) Reference; (c) FPN/Volume-br predictions; (d) Proposed (Multiscale) predictions
        \textcolor{OliveGreen}{\textbf{True positives}} are depicted in green;
        \textcolor{Blue}{\textbf{false positives}}, in blue; and \textcolor{BrickRed}{\textbf{false negatives}}, in red.
        This OCT volume is particularly challenging for RBV segmentation due to the presence of lesions resembling vessels (see blue false positives in (c)).
        These lesions are correctly identified as background by the proposed methods (bottom image) thanks to the complementary information provided by the SLO image.
    }
    \label{fig:qualitative_rbv_oct}
\end{figure}

Qualitative results on the test set of some of the methods under comparison
are shown in Fig.~\ref{fig:qualitative_rbv}.
Additionally, slice \#12 of the OCT corresponding to that sample, as well as the predictions for that slice of some of the models, are presented in Fig.~\ref{fig:qualitative_rbv_oct}.
As shown in the first row of Fig.~\ref{fig:qualitative_rbv}, the proposed fusion methods particularly contributed to improving the performance for cases with multiple pathological signs or lesions.
In this case, both Lachinov \textit{et al.} and FPN/Volume-br methods mistakenly identified lesion areas as vessels (in the figures, in blue).
Most of these mistakes were avoided by the proposed methods, which were able to correctly identify these regions as background.
This can be explained by the fact that these lesions resemble vessels in OCT (small bright dots dropping shadows), but not in SLO, so the fusion methods were able to take advantage of the complementary information provided by the SLO to correctly distinguish these structures.

\section{Conclusions}%
\label{sec:conclusions}

In this work, we have proposed a novel framework for fusing multimodal data of heterogeneous dimensionality that, in contrast to state-of-the-art methods, is compatible with localization tasks.
The proposed framework, based on the idea of LL fusion, aims to project the features extracted from the different modalities into a common feature subspace so that they can be fused and processed straightforwardly for obtaining the final prediction.
To validate the framework, we proposed and validated two different fusion approaches, Late Fusion and Multiscale Fusion, for the tasks of GA and RBV segmentation in multimodal retinal images.
Results show that the proposed fusion approaches outperformed monomodal state-of-the-art methods on both tasks, and that multimodal models were able to take advantage of the complementary information provided by the different modalities to give rise to more robust and accurate predictions.
The improvement was found particularly significant for the task of GA segmentation, where the proposed models outperformed the monomodal state of the art by a large margin.

As a general framework, the proposed method can be applied to any task involving multimodal data of heterogeneous dimensionality.
This includes other imaging modalities such as OCT-A and color fundus photography, or CT and X-ray.
Moreover, the proposed framework can be easily extended to the case of more than two modalities, by simply adding more branches to the network, and to multimodal data of the same dimensionality, by simply removing the projection layers and fusing the features extracted from the different modalities.

Despite the promising results, the proposed framework also presents some limitations.
The most important is the need for registration of the different modalities.
Although, in some cases, this registration is performed automatically by the imaging device (e.g., for OCT and SLO), in other cases it needs to be performed manually either by a clinician, a trained operator, or an external registration algorithm (e.g., for OCT and FAF).
This limitation represents an interesting line of future work,
which would ideally lead to the development of an automatic registration mechanism that could be integrated into the proposed framework.

In conclusion, the proposed framework and its implementations (which we have made publicly available) represent an effective approach to multimodal medical image analysis that can be applied to a wide range of tasks and modalities.
We believe that this work will encourage further research in this direction, and that it will contribute to the development of more robust and accurate multimodal medical image analysis systems, with a particular focus on the field of ophthalmology.

\section*{References}

\bibliographystyle{IEEEtran}
\bibliography{bibliography}

\end{document}